\documentclass[11pt]{article}
\usepackage[margin=1in]{geometry}
\usepackage[T1]{fontenc}
\usepackage{lmodern}
\usepackage{booktabs}
\usepackage{array}
\usepackage{graphicx}
\usepackage{amsmath}
\usepackage{url}
\usepackage[hidelinks]{hyperref}
\usepackage{caption}
\usepackage{placeins}
\newcommand{\BenchCalib}{635}
\newcommand{\BenchDev}{636}
\newcommand{\BenchTest}{463}
\newcommand{\BenchTrain}{4,925}
\newcommand{\BxAfterRho}{1.0}
\newcommand{\BxCap}{9,552}
\newcommand{\BxCross}{21,576}
\newcommand{\BxEq}{82.7}
\newcommand{\BxFlipfrac}{0.77}
\newcommand{\BxFlips}{17}
\newcommand{\BxMax}{0.992}
\newcommand{\BxN}{8,856}
\newcommand{\BxOld}{15,452}
\newcommand{\BxOldonly}{5,900}
\newcommand{\BxOoFict}{4,890}
\newcommand{\BxOoInel}{880}
\newcommand{\BxOoMisc}{130}
\newcommand{\BxOtherDwMax}{0.00271}
\newcommand{\BxOtherRhoMin}{0.999504}
\newcommand{\BxPairs}{2,214}
\newcommand{\BxRest}{12,024}
\newcommand{\BxRho}{0.990383}
\newcommand{\BxThrDw}{0.005}
\newcommand{\BxThrRho}{0.999}
\newcommand{\BxTok}{8,192}
\newcommand{\CvTrainChecked}{6,944}
\newcommand{\CvTrainClean}{6,196}
\newcommand{\CvTrainPct}{89.2}
\newcommand{\CvTrainVThree}{172}
\newcommand{\CvTrainVSix}{487}
\newcommand{\CvValChecked}{504}
\newcommand{\CvValClean}{463}
\newcommand{\CvValPct}{91.9}
\newcommand{\CvValVThree}{30}
\newcommand{\CxAboneCs}{0.000}
\newcommand{\CxAboneEc}{+0.021}
\newcommand{\CxAbthreeCs}{0.500}
\newcommand{\CxAbthreeEc}{+0.000}
\newcommand{\CxAbtwoCs}{0.000}
\newcommand{\CxAbtwoEc}{+0.495}
\newcommand{\CxAbtwoNom}{0.997}
\newcommand{\CxAbzeroCs}{0.000}
\newcommand{\CxAbzeroEc}{$-$0.022}
\newcommand{\CxAgreeCone}{11}
\newcommand{\CxAgreeCtwo}{11}
\newcommand{\CxAudGemmaNaturalL}{$-$0.604}
\newcommand{\CxAudLlamaNaturalL}{$-$0.499}
\newcommand{\CxAudPhiNaturalL}{$-$0.536}
\newcommand{\CxAudZyMax}{0.011}
\newcommand{\CxBootB}{10,000}
\newcommand{\CxCsLoMin}{0.290}
\newcommand{\CxCsMax}{0.484}
\newcommand{\CxCsMin}{0.308}
\newcommand{\CxFamM}{11}
\newcommand{\CxGemmaCone}{0.435}
\newcommand{\CxGemmaCs}{0.441}
\newcommand{\CxGemmaCsHi}{0.451}
\newcommand{\CxGemmaCsLo}{0.430}
\newcommand{\CxGemmaCtwo}{0.457}
\newcommand{\CxGemmaEc}{+0.197}
\newcommand{\CxGemmaEcHi}{+0.223}
\newcommand{\CxGemmaEcLo}{+0.170}
\newcommand{\CxGemmaPt}{$-$0.192}
\newcommand{\CxGemmaPtHi}{$-$0.164}
\newcommand{\CxGemmaPtLo}{$-$0.219}
\newcommand{\CxJudgeCone}{0.460}
\newcommand{\CxJudgeCs}{0.472}
\newcommand{\CxJudgeCsHi}{0.480}
\newcommand{\CxJudgeCsLo}{0.464}
\newcommand{\CxJudgeCtwo}{0.486}
\newcommand{\CxJudgeEc}{+0.268}
\newcommand{\CxJudgeEcHi}{+0.293}
\newcommand{\CxJudgeEcLo}{+0.242}
\newcommand{\CxJudgePt}{$-$0.215}
\newcommand{\CxJudgePtHi}{$-$0.189}
\newcommand{\CxJudgePtLo}{$-$0.242}
\newcommand{\CxLlamaCone}{0.330}
\newcommand{\CxLlamaCs}{0.308}
\newcommand{\CxLlamaCsHi}{0.326}
\newcommand{\CxLlamaCsLo}{0.290}
\newcommand{\CxLlamaCtwo}{0.379}
\newcommand{\CxLlamaEc}{+0.238}
\newcommand{\CxLlamaEcHi}{+0.265}
\newcommand{\CxLlamaEcLo}{+0.212}
\newcommand{\CxLlamaPt}{$-$0.192}
\newcommand{\CxLlamaPtHi}{$-$0.165}
\newcommand{\CxLlamaPtLo}{$-$0.221}
\newcommand{\CxMZeroCone}{0.380}
\newcommand{\CxMZeroCs}{0.407}
\newcommand{\CxMZeroCsHi}{0.417}
\newcommand{\CxMZeroCsLo}{0.396}
\newcommand{\CxMZeroCtwo}{0.451}
\newcommand{\CxMZeroEc}{+0.374}
\newcommand{\CxMZeroEcHi}{+0.394}
\newcommand{\CxMZeroEcLo}{+0.354}
\newcommand{\CxMZeroPt}{+0.171}
\newcommand{\CxMZeroPtHi}{+0.198}
\newcommand{\CxMZeroPtLo}{+0.142}
\newcommand{\CxMOneCone}{0.485}
\newcommand{\CxMOneCs}{0.481}
\newcommand{\CxMOneCsHi}{0.486}
\newcommand{\CxMOneCsLo}{0.476}
\newcommand{\CxMOneCtwo}{0.486}
\newcommand{\CxMOneEc}{+0.352}
\newcommand{\CxMOneEcHi}{+0.371}
\newcommand{\CxMOneEcLo}{+0.330}
\newcommand{\CxMOnePt}{$-$0.057}
\newcommand{\CxMOnePtHi}{$-$0.031}
\newcommand{\CxMOnePtLo}{$-$0.087}
\newcommand{\CxMTwoCone}{0.440}
\newcommand{\CxMTwoCs}{0.454}
\newcommand{\CxMTwoCsHi}{0.461}
\newcommand{\CxMTwoCsLo}{0.445}
\newcommand{\CxMTwoCtwo}{0.464}
\newcommand{\CxMTwoEc}{+0.300}
\newcommand{\CxMTwoEcHi}{+0.329}
\newcommand{\CxMTwoEcLo}{+0.286}
\newcommand{\CxMTwoPt}{+0.189}
\newcommand{\CxMTwoPtHi}{+0.215}
\newcommand{\CxMTwoPtLo}{+0.164}
\newcommand{\CxMThreeCone}{0.450}
\newcommand{\CxMThreeCs}{0.450}
\newcommand{\CxMThreeCsHi}{0.459}
\newcommand{\CxMThreeCsLo}{0.443}
\newcommand{\CxMThreeCtwo}{0.462}
\newcommand{\CxMThreeEc}{+0.240}
\newcommand{\CxMThreeEcHi}{+0.260}
\newcommand{\CxMThreeEcLo}{+0.214}
\newcommand{\CxMThreePt}{+0.133}
\newcommand{\CxMThreePtHi}{+0.166}
\newcommand{\CxMThreePtLo}{+0.114}
\newcommand{\CxMFoureCone}{0.485}
\newcommand{\CxMFoureCs}{0.484}
\newcommand{\CxMFoureCsHi}{0.489}
\newcommand{\CxMFoureCsLo}{0.478}
\newcommand{\CxMFoureCtwo}{0.474}
\newcommand{\CxMFoureEc}{+0.327}
\newcommand{\CxMFoureEcHi}{+0.349}
\newcommand{\CxMFoureEcLo}{+0.303}
\newcommand{\CxMFourePt}{$-$0.006}
\newcommand{\CxMFourePtHi}{+0.024}
\newcommand{\CxMFourePtLo}{$-$0.036}
\newcommand{\CxMFourjCone}{0.490}
\newcommand{\CxMFourjCs}{0.481}
\newcommand{\CxMFourjCsHi}{0.487}
\newcommand{\CxMFourjCsLo}{0.475}
\newcommand{\CxMFourjCtwo}{0.480}
\newcommand{\CxMFourjEc}{+0.157}
\newcommand{\CxMFourjEcHi}{+0.186}
\newcommand{\CxMFourjEcLo}{+0.128}
\newcommand{\CxMFourjPt}{+0.165}
\newcommand{\CxMFourjPtHi}{+0.192}
\newcommand{\CxMFourjPtLo}{+0.136}
\newcommand{\CxNCone}{50}
\newcommand{\CxNCs}{11}
\newcommand{\CxNCtwo}{231}
\newcommand{\CxNE}{2,214}
\newcommand{\CxNEc}{11}
\newcommand{\CxNPt}{10}
\newcommand{\CxNomMax}{0.998}
\newcommand{\CxNomMin}{0.732}
\newcommand{\CxPhiCone}{0.345}
\newcommand{\CxPhiCs}{0.379}
\newcommand{\CxPhiCsHi}{0.393}
\newcommand{\CxPhiCsLo}{0.365}
\newcommand{\CxPhiCtwo}{0.442}
\newcommand{\CxPhiEc}{+0.284}
\newcommand{\CxPhiEcHi}{+0.308}
\newcommand{\CxPhiEcLo}{+0.259}
\newcommand{\CxPhiPt}{$-$0.144}
\newcommand{\CxPhiPtHi}{$-$0.115}
\newcommand{\CxPhiPtLo}{$-$0.174}
\newcommand{\CxQwenCone}{0.395}
\newcommand{\CxQwenCs}{0.440}
\newcommand{\CxQwenCsHi}{0.451}
\newcommand{\CxQwenCsLo}{0.428}
\newcommand{\CxQwenCtwo}{0.465}
\newcommand{\CxQwenEc}{+0.240}
\newcommand{\CxQwenEcHi}{+0.266}
\newcommand{\CxQwenEcLo}{+0.213}
\newcommand{\CxQwenPt}{$-$0.181}
\newcommand{\CxQwenPtHi}{$-$0.152}
\newcommand{\CxQwenPtLo}{$-$0.209}
\newcommand{\FeOneMuFictJDce}{+0.041}
\newcommand{\FeOneMuFictJDceChi}{+0.094}
\newcommand{\FeOneMuFictJDceClo}{+0.000}
\newcommand{\FeOneMuFictJDcePhi}{+0.082}
\newcommand{\FeOneMuFictJDcePlo}{+0.010}
\newcommand{\FeOneMuFictJSigm}{+0.480}
\newcommand{\FeOneMuFictKeff}{14.64}
\newcommand{\FeOneMuFictLargest}{20.4}
\newcommand{\FeOneMuFictMzeroCem}{+0.306}
\newcommand{\FeOneMuFictMzeroCeo}{$-$0.276}
\newcommand{\FeOneMuFictMzeroSigm}{$-$0.173}
\newcommand{\FeOneMuFictMzeroSigmChi}{$-$0.056}
\newcommand{\FeOneMuFictMzeroSigmClo}{$-$0.259}
\newcommand{\FeOneMuFictMzeroSigo}{$-$0.745}
\newcommand{\FeOneMuFictNcl}{40}
\newcommand{\FeOneMuFictNq}{98}
\newcommand{\FeOneMuFictRho}{0.829}
\newcommand{\FeOneMuJointKeff}{18.22}
\newcommand{\FeOneMuJointLargest}{18.2}
\newcommand{\FeOneMuJointNcl}{52}
\newcommand{\FeOneMuJointNq}{220}
\newcommand{\FeOneMuRealJDce}{+0.107}
\newcommand{\FeOneMuRealJDceChi}{+0.214}
\newcommand{\FeOneMuRealJDceClo}{+0.025}
\newcommand{\FeOneMuRealJDcePhi}{+0.180}
\newcommand{\FeOneMuRealJDcePlo}{+0.033}
\newcommand{\FeOneMuRealJSigm}{+0.246}
\newcommand{\FeOneMuRealJSigmChi}{+0.393}
\newcommand{\FeOneMuRealJSigmClo}{+0.082}
\newcommand{\FeOneMuRealJSigmPhi}{+0.344}
\newcommand{\FeOneMuRealJSigmPlo}{+0.147}
\newcommand{\FeOneMuRealKeff}{20.9}
\newcommand{\FeOneMuRealLargest}{16.4}
\newcommand{\FeOneMuRealMzeroCem}{+0.180}
\newcommand{\FeOneMuRealMzeroCeo}{$-$0.205}
\newcommand{\FeOneMuRealMzeroSigm}{$-$0.279}
\newcommand{\FeOneMuRealMzeroSigmChi}{$-$0.190}
\newcommand{\FeOneMuRealMzeroSigmClo}{$-$0.358}
\newcommand{\FeOneMuRealMzeroSigo}{$-$0.656}
\newcommand{\FeOneMuRealNcl}{52}
\newcommand{\FeOneMuRealNq}{122}
\newcommand{\FeOneMuRealRho}{0.657}
\newcommand{\FeOneWikiFictJDce}{+0.014}
\newcommand{\FeOneWikiFictJDcePhi}{+0.050}
\newcommand{\FeOneWikiFictJDcePlo}{$-$0.021}
\newcommand{\FeOneWikiFictJSigm}{+0.418}
\newcommand{\FeOneWikiFictKeff}{1.63}
\newcommand{\FeOneWikiFictLargest}{78.0}
\newcommand{\FeOneWikiFictNcl}{19}
\newcommand{\FeOneWikiFictNq}{141}
\newcommand{\FeOneWikiFictRho}{0.943}
\newcommand{\FeOneWikiJointKeff}{1.68}
\newcommand{\FeOneWikiJointLargest}{76.9}
\newcommand{\FeOneWikiJointNcl}{28}
\newcommand{\FeOneWikiJointNq}{428}
\newcommand{\FeOneWikiRealJDce}{+0.129}
\newcommand{\FeOneWikiRealJDceChi}{+0.177}
\newcommand{\FeOneWikiRealJDceClo}{$-$0.020}
\newcommand{\FeOneWikiRealJDcePhi}{+0.171}
\newcommand{\FeOneWikiRealJDcePlo}{+0.087}
\newcommand{\FeOneWikiRealJSigm}{+0.289}
\newcommand{\FeOneWikiRealKeff}{1.7}
\newcommand{\FeOneWikiRealLargest}{76.3}
\newcommand{\FeOneWikiRealMzeroSigm}{+0.066}
\newcommand{\FeOneWikiRealMzeroSigmChi}{+0.109}
\newcommand{\FeOneWikiRealMzeroSigmClo}{$-$0.053}
\newcommand{\FeOneWikiRealMzeroSigo}{$-$0.084}
\newcommand{\FeOneWikiRealNcl}{28}
\newcommand{\FeOneWikiRealNq}{287}
\newcommand{\FeOneWikiRealRho}{1.000}
\newcommand{\FeTwoMargin}{0.1}
\newcommand{\FeTwoMuFictJDmask}{$-$0.031}
\newcommand{\FeTwoMuFictKeff}{14.36}
\newcommand{\FeTwoMuFictLargest}{20.6}
\newcommand{\FeTwoMuFictNcl}{39}
\newcommand{\FeTwoMuFictNhard}{98}
\newcommand{\FeTwoMuFictNkept}{97}
\newcommand{\FeTwoMuFictNq}{97}
\newcommand{\FeTwoMuJointKeff}{17.52}
\newcommand{\FeTwoMuJointLargest}{18.6}
\newcommand{\FeTwoMuJointNcl}{51}
\newcommand{\FeTwoMuJointNq}{215}
\newcommand{\FeTwoMuRealAXintitleu}{20.3}
\newcommand{\FeTwoMuRealDTokm}{1,238}
\newcommand{\FeTwoMuRealDToku}{1,261}
\newcommand{\FeTwoMuRealJCeo}{+0.339}
\newcommand{\FeTwoMuRealJCeu}{+0.356}
\newcommand{\FeTwoMuRealJDmask}{$-$0.025}
\newcommand{\FeTwoMuRealKeff}{19.89}
\newcommand{\FeTwoMuRealLargest}{17.0}
\newcommand{\FeTwoMuRealMthreeDmask}{+0.072}
\newcommand{\FeTwoMuRealMzeroDmask}{+0.009}
\newcommand{\FeTwoMuRealNcl}{51}
\newcommand{\FeTwoMuRealNhard}{122}
\newcommand{\FeTwoMuRealNkept}{118}
\newcommand{\FeTwoMuRealNq}{118}
\newcommand{\FeTwoWikiFictJDce}{+0.000}
\newcommand{\FeTwoWikiFictJDcePhi}{+0.043}
\newcommand{\FeTwoWikiFictJDcePlo}{$-$0.043}
\newcommand{\FeTwoWikiFictJDmask}{$-$0.014}
\newcommand{\FeTwoWikiFictJDmaskChi}{+0.080}
\newcommand{\FeTwoWikiFictJDmaskClo}{$-$0.127}
\newcommand{\FeTwoWikiFictJDmaskPhi}{+0.029}
\newcommand{\FeTwoWikiFictJDmaskPlo}{$-$0.057}
\newcommand{\FeTwoWikiFictJSigm}{+0.393}
\newcommand{\FeTwoWikiFictKeff}{1.64}
\newcommand{\FeTwoWikiFictLargest}{77.9}
\newcommand{\FeTwoWikiFictNcl}{19}
\newcommand{\FeTwoWikiFictNhard}{141}
\newcommand{\FeTwoWikiFictNkept}{140}
\newcommand{\FeTwoWikiFictNq}{140}
\newcommand{\FeTwoWikiFictRhom}{1.000}
\newcommand{\FeTwoWikiJointKeff}{1.68}
\newcommand{\FeTwoWikiJointLargest}{76.8}
\newcommand{\FeTwoWikiJointNcl}{29}
\newcommand{\FeTwoWikiJointNq}{418}
\newcommand{\FeTwoWikiRealAXintitleu}{100.0}
\newcommand{\FeTwoWikiRealDTokm}{760}
\newcommand{\FeTwoWikiRealDToku}{784}
\newcommand{\FeTwoWikiRealDXinbodym}{100.0}
\newcommand{\FeTwoWikiRealDYintitlem}{0.0}
\newcommand{\FeTwoWikiRealDYintitleu}{100.0}
\newcommand{\FeTwoWikiRealJDce}{$-$0.018}
\newcommand{\FeTwoWikiRealJDcePhi}{+0.022}
\newcommand{\FeTwoWikiRealJDcePlo}{$-$0.058}
\newcommand{\FeTwoWikiRealJDmask}{$-$0.018}
\newcommand{\FeTwoWikiRealJDmaskChi}{+0.078}
\newcommand{\FeTwoWikiRealJDmaskClo}{$-$0.042}
\newcommand{\FeTwoWikiRealJDmaskPhi}{+0.022}
\newcommand{\FeTwoWikiRealJDmaskPlo}{$-$0.054}
\newcommand{\FeTwoWikiRealJSigm}{+0.180}
\newcommand{\FeTwoWikiRealKeff}{1.7}
\newcommand{\FeTwoWikiRealLargest}{76.3}
\newcommand{\FeTwoWikiRealMthreeDmask}{$-$0.045}
\newcommand{\FeTwoWikiRealMthreeDmaskPhi}{$-$0.014}
\newcommand{\FeTwoWikiRealMthreeDmaskPlo}{$-$0.072}
\newcommand{\FeTwoWikiRealMzeroDmask}{$-$0.061}
\newcommand{\FeTwoWikiRealMzeroDmaskPhi}{$-$0.029}
\newcommand{\FeTwoWikiRealMzeroDmaskPlo}{$-$0.095}
\newcommand{\FeTwoWikiRealNcl}{29}
\newcommand{\FeTwoWikiRealNhard}{295}
\newcommand{\FeTwoWikiRealNkept}{278}
\newcommand{\FeTwoWikiRealNq}{278}
\newcommand{\FeTwoWikiRealRhom}{1.000}
\newcommand{\FinExposedPct}{3.5}
\newcommand{\FinGbDelta}{0.022}
\newcommand{\FinGbHi}{0.038}
\newcommand{\FinGbLo}{0.007}
\newcommand{\FinGbMargin}{0.05}
\newcommand{\FinGcDelta}{0.000}
\newcommand{\FinGcHi}{0.009}
\newcommand{\FinGcLo}{$-$0.009}
\newcommand{\FinNHard}{447}
\newcommand{\FinNomMZero}{0.970}
\newcommand{\FinNomMThree}{0.990}
\newcommand{\FinNomMFoure}{0.996}
\newcommand{\FinNomJudge}{0.893}
\newcommand{\GateGZeroMax}{0.65}
\newcommand{\GeMZero}{0.456}
\newcommand{\GeMZerolast}{0.534}
\newcommand{\GeMOne}{0.421}
\newcommand{\GeMTwo}{0.148}
\newcommand{\GeMThree}{0.150}
\newcommand{\GeMFoure}{0.076}
\newcommand{\GeMFourj}{0.066}
\newcommand{\GeJudge}{0.168}
\newcommand{\GrpDisjointPct}{58.9}
\newcommand{\GrpGZeroBeforeFix}{0.608}
\newcommand{\GrpGroupsAfter}{18,534}
\newcommand{\GrpGroupsBefore}{120}
\newcommand{\GrpMaxGroup}{48}
\newcommand{\GrpSharedDocs}{196}
\newcommand{\GrpSingletonPct}{88.3}
\newcommand{\GrpTrainDocs}{12,304}
\newcommand{\GrpValDisjoint}{1,424}
\newcommand{\GrpValDocs}{2,328}
\newcommand{\GrpValTotal}{2,417}
\newcommand{\HpAfter}{2,453}
\newcommand{\HpBefore}{2,650}
\newcommand{\HpGemmaCe}{+0.322}
\newcommand{\HpGemmaEe}{+0.046}
\newcommand{\HpGemmaNom}{0.859}
\newcommand{\HpGemmaSig}{+0.276}
\newcommand{\HpGemmaSigHi}{+0.294}
\newcommand{\HpGemmaSigLo}{+0.257}
\newcommand{\HpGemmaSigm}{+0.399}
\newcommand{\HpGemmaSigmHi}{+0.413}
\newcommand{\HpGemmaSigmLo}{+0.386}
\newcommand{\HpHOneN}{6}
\newcommand{\HpHOnePass}{5}
\newcommand{\HpJudgeCe}{+0.321}
\newcommand{\HpJudgeEe}{+0.303}
\newcommand{\HpJudgeNom}{0.969}
\newcommand{\HpJudgeSig}{+0.018}
\newcommand{\HpJudgeSigHi}{+0.042}
\newcommand{\HpJudgeSigLo}{$-$0.007}
\newcommand{\HpJudgeSiglo}{$-$0.007}
\newcommand{\HpJudgeSigm}{+0.171}
\newcommand{\HpJudgeSigmHi}{+0.189}
\newcommand{\HpJudgeSigmLo}{+0.153}
\newcommand{\HpKeff}{2,276}
\newcommand{\HpLlamaCe}{+0.140}
\newcommand{\HpLlamaEe}{+0.080}
\newcommand{\HpLlamaNom}{0.729}
\newcommand{\HpLlamaSig}{+0.059}
\newcommand{\HpLlamaSigHi}{+0.091}
\newcommand{\HpLlamaSigLo}{+0.025}
\newcommand{\HpLlamaSigm}{+0.242}
\newcommand{\HpLlamaSigmHi}{+0.267}
\newcommand{\HpLlamaSigmLo}{+0.215}
\newcommand{\HpMZeroCe}{+0.192}
\newcommand{\HpMZeroEe}{+0.432}
\newcommand{\HpMZeroNom}{0.995}
\newcommand{\HpMZeroSig}{$-$0.240}
\newcommand{\HpMZeroSigHi}{$-$0.216}
\newcommand{\HpMZeroSigLo}{$-$0.262}
\newcommand{\HpMZeroSigm}{$-$0.019}
\newcommand{\HpMZeroSigmHi}{$-$0.003}
\newcommand{\HpMZeroSigmLo}{$-$0.035}
\newcommand{\HpMZerolastCe}{+0.312}
\newcommand{\HpMZerolastEe}{+0.324}
\newcommand{\HpMZerolastNom}{0.975}
\newcommand{\HpMZerolastSig}{$-$0.012}
\newcommand{\HpMZerolastSigHi}{+0.013}
\newcommand{\HpMZerolastSigLo}{$-$0.036}
\newcommand{\HpMZerolastSigm}{+0.132}
\newcommand{\HpMZerolastSigmHi}{+0.150}
\newcommand{\HpMZerolastSigmLo}{+0.114}
\newcommand{\HpMOneCe}{+0.355}
\newcommand{\HpMOneEe}{+0.358}
\newcommand{\HpMOneNom}{0.996}
\newcommand{\HpMOneSig}{$-$0.004}
\newcommand{\HpMOneSigHi}{+0.022}
\newcommand{\HpMOneSigLo}{$-$0.021}
\newcommand{\HpMOneSigm}{+0.125}
\newcommand{\HpMOneSigmHi}{+0.145}
\newcommand{\HpMOneSigmLo}{+0.116}
\newcommand{\HpMTwoCe}{+0.347}
\newcommand{\HpMTwoEe}{+0.456}
\newcommand{\HpMTwoNom}{0.996}
\newcommand{\HpMTwoSig}{$-$0.108}
\newcommand{\HpMTwoSigHi}{$-$0.097}
\newcommand{\HpMTwoSigLo}{$-$0.130}
\newcommand{\HpMTwoSigm}{$-$0.002}
\newcommand{\HpMTwoSigmHi}{+0.009}
\newcommand{\HpMTwoSigmLo}{$-$0.012}
\newcommand{\HpMThreeCe}{+0.365}
\newcommand{\HpMThreeEe}{+0.436}
\newcommand{\HpMThreeNom}{0.994}
\newcommand{\HpMThreeSig}{$-$0.071}
\newcommand{\HpMThreeSigHi}{$-$0.059}
\newcommand{\HpMThreeSigLo}{$-$0.090}
\newcommand{\HpMThreeSigm}{+0.022}
\newcommand{\HpMThreeSigmHi}{+0.031}
\newcommand{\HpMThreeSigmLo}{+0.009}
\newcommand{\HpMFoureCe}{+0.418}
\newcommand{\HpMFoureEe}{+0.470}
\newcommand{\HpMFoureNom}{0.997}
\newcommand{\HpMFoureSig}{$-$0.052}
\newcommand{\HpMFoureSigHi}{$-$0.038}
\newcommand{\HpMFoureSigLo}{$-$0.066}
\newcommand{\HpMFoureSigm}{+0.011}
\newcommand{\HpMFoureSigmHi}{+0.020}
\newcommand{\HpMFoureSigmLo}{+0.002}
\newcommand{\HpMFourjCe}{+0.456}
\newcommand{\HpMFourjEe}{+0.474}
\newcommand{\HpMFourjNom}{0.998}
\newcommand{\HpMFourjSig}{$-$0.018}
\newcommand{\HpMFourjSigHi}{$-$0.008}
\newcommand{\HpMFourjSigLo}{$-$0.029}
\newcommand{\HpMFourjSigm}{+0.007}
\newcommand{\HpMFourjSigmHi}{+0.015}
\newcommand{\HpMFourjSigmLo}{$-$0.002}
\newcommand{\HpMoneSighi}{+0.022}
\newcommand{\HpN}{2,276}
\newcommand{\HpPhiCe}{+0.182}
\newcommand{\HpPhiEe}{+0.210}
\newcommand{\HpPhiNom}{0.856}
\newcommand{\HpPhiSig}{$-$0.027}
\newcommand{\HpPhiSigHi}{+0.004}
\newcommand{\HpPhiSigLo}{$-$0.057}
\newcommand{\HpPhiSiglo}{$-$0.057}
\newcommand{\HpPhiSigm}{+0.169}
\newcommand{\HpPhiSigmHi}{+0.193}
\newcommand{\HpPhiSigmLo}{+0.145}
\newcommand{\HpQwenCe}{+0.292}
\newcommand{\HpQwenEe}{+0.167}
\newcommand{\HpQwenNom}{0.913}
\newcommand{\HpQwenSig}{+0.126}
\newcommand{\HpQwenSigHi}{+0.153}
\newcommand{\HpQwenSigLo}{+0.098}
\newcommand{\HpQwenSigm}{+0.269}
\newcommand{\HpQwenSigmHi}{+0.290}
\newcommand{\HpQwenSigmLo}{+0.246}
\newcommand{\HpSigmMax}{+0.125}
\newcommand{\HpSigmMin}{$-$0.019}
\newcommand{\HpXjN}{4}
\newcommand{\HpXjPass}{3}
\newcommand{\MZeroBestAuc}{0.993}
\newcommand{\MZeroBestLayer}{36}
\newcommand{\MFourLoraParams}{5,242,880}
\newcommand{\MFourTrainPairs}{4,925}
\newcommand{\MuDExtraJudgesN}{4}
\newcommand{\MuDExtraJudgesPass}{1}
\newcommand{\MuFictMZeroCe}{$-$0.276}
\newcommand{\MuFictMZeroEe}{+0.469}
\newcommand{\MuFictMZeroNom}{0.929}
\newcommand{\MuFictMZeroSig}{$-$0.745}
\newcommand{\MuFictMZeroSigHi}{$-$0.643}
\newcommand{\MuFictMZeroSigLo}{$-$0.832}
\newcommand{\MuFictMZerolastCe}{+0.010}
\newcommand{\MuFictMZerolastEe}{+0.398}
\newcommand{\MuFictMZerolastNom}{0.918}
\newcommand{\MuFictMZerolastSig}{$-$0.388}
\newcommand{\MuFictMZerolastSigHi}{$-$0.255}
\newcommand{\MuFictMZerolastSigLo}{$-$0.515}
\newcommand{\MuFictMOneCe}{+0.112}
\newcommand{\MuFictMOneEe}{+0.306}
\newcommand{\MuFictMOneNom}{0.888}
\newcommand{\MuFictMOneSig}{$-$0.194}
\newcommand{\MuFictMOneSigHi}{$-$0.027}
\newcommand{\MuFictMOneSigLo}{$-$0.296}
\newcommand{\MuFictMTwoCe}{+0.112}
\newcommand{\MuFictMTwoEe}{+0.480}
\newcommand{\MuFictMTwoNom}{0.929}
\newcommand{\MuFictMTwoSig}{$-$0.367}
\newcommand{\MuFictMTwoSigHi}{$-$0.296}
\newcommand{\MuFictMTwoSigLo}{$-$0.480}
\newcommand{\MuFictMThreeCe}{+0.173}
\newcommand{\MuFictMThreeEe}{+0.459}
\newcommand{\MuFictMThreeNom}{0.929}
\newcommand{\MuFictMThreeSig}{$-$0.286}
\newcommand{\MuFictMThreeSigHi}{$-$0.194}
\newcommand{\MuFictMThreeSigLo}{$-$0.374}
\newcommand{\MuFictMFoureCe}{+0.327}
\newcommand{\MuFictMFoureEe}{+0.500}
\newcommand{\MuFictMFoureNom}{0.908}
\newcommand{\MuFictMFoureSig}{$-$0.173}
\newcommand{\MuFictMFoureSigHi}{$-$0.102}
\newcommand{\MuFictMFoureSigLo}{$-$0.255}
\newcommand{\MuFictMFourjCe}{+0.337}
\newcommand{\MuFictMFourjEe}{+0.490}
\newcommand{\MuFictMFourjNom}{1.000}
\newcommand{\MuFictMFourjSig}{$-$0.153}
\newcommand{\MuFictMFourjSigHi}{$-$0.082}
\newcommand{\MuFictMFourjSigLo}{$-$0.235}
\newcommand{\MuFictGemmaCe}{+0.403}
\newcommand{\MuFictGemmaEe}{$-$0.143}
\newcommand{\MuFictGemmaNom}{0.862}
\newcommand{\MuFictGemmaSig}{+0.260}
\newcommand{\MuFictGemmaSigHi}{0.342}
\newcommand{\MuFictGemmaSigLo}{0.184}
\newcommand{\MuFictJudgeCe}{+0.459}
\newcommand{\MuFictJudgeEe}{$-$0.031}
\newcommand{\MuFictJudgeNom}{0.929}
\newcommand{\MuFictJudgeSig}{+0.429}
\newcommand{\MuFictJudgeSigHi}{0.469}
\newcommand{\MuFictJudgeSigLo}{0.337}
\newcommand{\MuFictLlamaCe}{+0.214}
\newcommand{\MuFictLlamaEe}{$-$0.031}
\newcommand{\MuFictLlamaNom}{0.725}
\newcommand{\MuFictLlamaSig}{+0.184}
\newcommand{\MuFictLlamaSigHi}{0.255}
\newcommand{\MuFictLlamaSigLo}{0.061}
\newcommand{\MuFictNhard}{98}
\newcommand{\MuFictPhiCe}{+0.122}
\newcommand{\MuFictPhiEe}{+0.122}
\newcommand{\MuFictPhiNom}{0.775}
\newcommand{\MuFictPhiSig}{+0.000}
\newcommand{\MuFictPhiSigHi}{0.153}
\newcommand{\MuFictPhiSigLo}{$-$0.163}
\newcommand{\MuFictQwenCe}{+0.347}
\newcommand{\MuFictQwenEe}{$-$0.133}
\newcommand{\MuFictQwenNom}{0.775}
\newcommand{\MuFictQwenSig}{+0.214}
\newcommand{\MuFictQwenSigHi}{0.306}
\newcommand{\MuFictQwenSigLo}{0.122}
\newcommand{\MuHThree}{0.219}
\newcommand{\MuHThreeHi}{0.344}
\newcommand{\MuHThreeLo}{0.092}
\newcommand{\MuRealMZeroCe}{$-$0.205}
\newcommand{\MuRealMZeroEe}{+0.451}
\newcommand{\MuRealMZeroNom}{0.934}
\newcommand{\MuRealMZeroSig}{$-$0.656}
\newcommand{\MuRealMZeroSigHi}{$-$0.553}
\newcommand{\MuRealMZeroSigLo}{$-$0.750}
\newcommand{\MuRealMZerolastCe}{$-$0.074}
\newcommand{\MuRealMZerolastEe}{+0.377}
\newcommand{\MuRealMZerolastNom}{0.902}
\newcommand{\MuRealMZerolastSig}{$-$0.451}
\newcommand{\MuRealMZerolastSigHi}{$-$0.328}
\newcommand{\MuRealMZerolastSigLo}{$-$0.561}
\newcommand{\MuRealMOneCe}{$-$0.016}
\newcommand{\MuRealMOneEe}{+0.410}
\newcommand{\MuRealMOneNom}{0.918}
\newcommand{\MuRealMOneSig}{$-$0.426}
\newcommand{\MuRealMOneSigHi}{$-$0.311}
\newcommand{\MuRealMOneSigLo}{$-$0.534}
\newcommand{\MuRealMTwoCe}{$-$0.057}
\newcommand{\MuRealMTwoEe}{+0.443}
\newcommand{\MuRealMTwoNom}{0.926}
\newcommand{\MuRealMTwoSig}{$-$0.500}
\newcommand{\MuRealMTwoSigHi}{$-$0.426}
\newcommand{\MuRealMTwoSigLo}{$-$0.603}
\newcommand{\MuRealMThreeCe}{+0.057}
\newcommand{\MuRealMThreeEe}{+0.467}
\newcommand{\MuRealMThreeNom}{0.926}
\newcommand{\MuRealMThreeSig}{$-$0.410}
\newcommand{\MuRealMThreeSigHi}{$-$0.303}
\newcommand{\MuRealMThreeSigLo}{$-$0.481}
\newcommand{\MuRealMFoureCe}{+0.271}
\newcommand{\MuRealMFoureEe}{+0.492}
\newcommand{\MuRealMFoureNom}{0.963}
\newcommand{\MuRealMFoureSig}{$-$0.221}
\newcommand{\MuRealMFoureSigHi}{$-$0.147}
\newcommand{\MuRealMFoureSigLo}{$-$0.303}
\newcommand{\MuRealMFourjCe}{+0.328}
\newcommand{\MuRealMFourjEe}{+0.475}
\newcommand{\MuRealMFourjNom}{1.000}
\newcommand{\MuRealMFourjSig}{$-$0.147}
\newcommand{\MuRealMFourjSigHi}{$-$0.074}
\newcommand{\MuRealMFourjSigLo}{$-$0.230}
\newcommand{\MuRealDJudgeRankNominal}{3}
\newcommand{\MuRealDJudgeRankSigma}{1}
\newcommand{\MuRealDNRankedSystems}{8}
\newcommand{\MuRealDNominalMax}{1.000}
\newcommand{\MuRealDNominalMin}{0.902}
\newcommand{\MuRealDTrainedEELoMax}{0.475}
\newcommand{\MuRealDTrainedEELoMin}{0.361}
\newcommand{\MuRealGemmaCe}{+0.414}
\newcommand{\MuRealGemmaEe}{$-$0.033}
\newcommand{\MuRealGemmaNom}{0.873}
\newcommand{\MuRealGemmaSig}{+0.381}
\newcommand{\MuRealGemmaSigHi}{0.434}
\newcommand{\MuRealGemmaSigLo}{0.299}
\newcommand{\MuRealJudgeCe}{+0.328}
\newcommand{\MuRealJudgeEe}{+0.189}
\newcommand{\MuRealJudgeNom}{0.943}
\newcommand{\MuRealJudgeSig}{+0.139}
\newcommand{\MuRealJudgeSigHi}{0.262}
\newcommand{\MuRealJudgeSigLo}{0.025}
\newcommand{\MuRealLlamaCe}{+0.147}
\newcommand{\MuRealLlamaEe}{+0.082}
\newcommand{\MuRealLlamaNom}{0.779}
\newcommand{\MuRealLlamaSig}{+0.066}
\newcommand{\MuRealLlamaSigHi}{0.197}
\newcommand{\MuRealLlamaSigLo}{$-$0.082}
\newcommand{\MuRealNhard}{122}
\newcommand{\MuRealPhiCe}{+0.049}
\newcommand{\MuRealPhiEe}{+0.254}
\newcommand{\MuRealPhiNom}{0.877}
\newcommand{\MuRealPhiSig}{$-$0.205}
\newcommand{\MuRealPhiSigHi}{$-$0.066}
\newcommand{\MuRealPhiSigLo}{$-$0.336}
\newcommand{\MuRealQwenCe}{+0.205}
\newcommand{\MuRealQwenEe}{+0.115}
\newcommand{\MuRealQwenNom}{0.820}
\newcommand{\MuRealQwenSig}{+0.090}
\newcommand{\MuRealQwenSigHi}{0.238}
\newcommand{\MuRealQwenSigLo}{$-$0.049}
\newcommand{\PhBootB}{10000}
\newcommand{\PhKeffMin}{20}
\newcommand{\PhMuClusterFictSigHi}{0.468}
\newcommand{\PhMuClusterFictSigLo}{0.216}
\newcommand{\PhMuClusterHThree}{0.219}
\newcommand{\PhMuClusterHThreeHi}{0.347}
\newcommand{\PhMuClusterHThreeLo}{0.095}
\newcommand{\PhMuClusterRealSigHi}{0.341}
\newcommand{\PhMuClusterRealSigLo}{$-$0.089}
\newcommand{\PhMuKeff}{18.22}
\newcommand{\PhMuLargest}{18.2}
\newcommand{\PhMuMZeroBitequal}{14}
\newcommand{\PhMuMZeroFlips}{4}
\newcommand{\PhMuMZeroNomLedger}{0.934}
\newcommand{\PhMuMZeroNomNow}{0.918}
\newcommand{\PhMuMZeroSingleNom}{0.934}
\newcommand{\PhMuMZeroTies}{10}
\newcommand{\PhMuNclusters}{52}
\newcommand{\PhMuNquads}{220}
\newcommand{\PhMuPairJm}{+0.590}
\newcommand{\PhMuPairJmHi}{0.738}
\newcommand{\PhMuPairJmLo}{0.443}
\newcommand{\PhMuPairRealSigHi}{0.262}
\newcommand{\PhMuPairRealSigLo}{0.016}
\newcommand{\PhMuTitleKeff}{2.03}
\newcommand{\PhPerVariantMin}{20}
\newcommand{\PhWikiClusterFictSigHi}{0.411}
\newcommand{\PhWikiClusterFictSigLo}{0.206}
\newcommand{\PhWikiClusterHThree}{0.118}
\newcommand{\PhWikiClusterHThreeHi}{0.260}
\newcommand{\PhWikiClusterHThreeLo}{0.068}
\newcommand{\PhWikiClusterRealSigHi}{0.255}
\newcommand{\PhWikiClusterRealSigLo}{0.054}
\newcommand{\PhWikiLargest}{76.1}
\newcommand{\PhWikiNclusters}{30}
\newcommand{\PhWikiNquads}{436}
\newcommand{\PhWikiPairJm}{+0.003}
\newcommand{\PhWikiPairJmHi}{0.091}
\newcommand{\PhWikiPairJmLo}{$-$0.083}
\newcommand{\PhWikiPairRealSigHi}{0.237}
\newcommand{\PhWikiPairRealSigLo}{0.088}
\newcommand{\SurfAbsentAuc}{0.959}
\newcommand{\SurfBmTwoFiveNonzeroPct}{2.38}
\newcommand{\SurfRemovalAuc}{0.922}
\newcommand{\SurfSubstTest}{0.568}
\newcommand{\TestFictMZeroCe}{$-$0.062}
\newcommand{\TestFictMZeroEe}{+0.464}
\newcommand{\TestFictMZeroNom}{0.991}
\newcommand{\TestFictMZeroSig}{$-$0.527}
\newcommand{\TestFictMZeroSigHi}{$-$0.451}
\newcommand{\TestFictMZeroSigLo}{$-$0.603}
\newcommand{\TestFictMZerolastCe}{+0.183}
\newcommand{\TestFictMZerolastEe}{+0.317}
\newcommand{\TestFictMZerolastNom}{0.942}
\newcommand{\TestFictMZerolastSig}{$-$0.134}
\newcommand{\TestFictMZerolastSigHi}{$-$0.045}
\newcommand{\TestFictMZerolastSigLo}{$-$0.228}
\newcommand{\TestFictMOneCe}{+0.268}
\newcommand{\TestFictMOneEe}{+0.299}
\newcommand{\TestFictMOneNom}{0.978}
\newcommand{\TestFictMOneSig}{$-$0.031}
\newcommand{\TestFictMOneSigHi}{0.039}
\newcommand{\TestFictMOneSigLo}{$-$0.125}
\newcommand{\TestFictMTwoCe}{+0.290}
\newcommand{\TestFictMTwoEe}{+0.451}
\newcommand{\TestFictMTwoNom}{0.991}
\newcommand{\TestFictMTwoSig}{$-$0.161}
\newcommand{\TestFictMTwoSigHi}{$-$0.119}
\newcommand{\TestFictMTwoSigLo}{$-$0.232}
\newcommand{\TestFictMThreeCe}{+0.317}
\newcommand{\TestFictMThreeEe}{+0.442}
\newcommand{\TestFictMThreeNom}{0.987}
\newcommand{\TestFictMThreeSig}{$-$0.125}
\newcommand{\TestFictMThreeSigHi}{$-$0.065}
\newcommand{\TestFictMThreeSigLo}{$-$0.176}
\newcommand{\TestFictMFoureCe}{+0.438}
\newcommand{\TestFictMFoureEe}{+0.491}
\newcommand{\TestFictMFoureNom}{1.000}
\newcommand{\TestFictMFoureSig}{$-$0.054}
\newcommand{\TestFictMFoureSigHi}{$-$0.022}
\newcommand{\TestFictMFoureSigLo}{$-$0.089}
\newcommand{\TestFictMFourjCe}{+0.438}
\newcommand{\TestFictMFourjEe}{+0.460}
\newcommand{\TestFictMFourjNom}{0.996}
\newcommand{\TestFictMFourjSig}{$-$0.022}
\newcommand{\TestFictMFourjSigHi}{0.018}
\newcommand{\TestFictMFourjSigLo}{$-$0.067}
\newcommand{\TestFictJudgeCe}{+0.406}
\newcommand{\TestFictJudgeEe}{+0.000}
\newcommand{\TestFictJudgeEeHi}{0.067}
\newcommand{\TestFictJudgeEeLo}{$-$0.067}
\newcommand{\TestFictJudgeNom}{0.920}
\newcommand{\TestFictJudgeSig}{+0.406}
\newcommand{\TestFictJudgeSigHi}{0.429}
\newcommand{\TestFictJudgeSigLo}{0.317}
\newcommand{\TestFictNcand}{270}
\newcommand{\TestFictNhard}{224}
\newcommand{\TestFictNvalid}{234}
\newcommand{\TestFictQEight}{36}
\newcommand{\TestRealMZeroCe}{$-$0.025}
\newcommand{\TestRealMZeroCeHi}{0.031}
\newcommand{\TestRealMZeroCeLo}{$-$0.080}
\newcommand{\TestRealMZeroEe}{+0.438}
\newcommand{\TestRealMZeroNom}{0.968}
\newcommand{\TestRealMZeroSig}{$-$0.463}
\newcommand{\TestRealMZeroSigHi}{$-$0.395}
\newcommand{\TestRealMZeroSigLo}{$-$0.528}
\newcommand{\TestRealMZerolastCe}{+0.120}
\newcommand{\TestRealMZerolastEe}{+0.265}
\newcommand{\TestRealMZerolastNom}{0.912}
\newcommand{\TestRealMZerolastSig}{$-$0.145}
\newcommand{\TestRealMZerolastSigHi}{$-$0.065}
\newcommand{\TestRealMZerolastSigLo}{$-$0.225}
\newcommand{\TestRealMOneCe}{+0.157}
\newcommand{\TestRealMOneEe}{+0.312}
\newcommand{\TestRealMOneNom}{0.971}
\newcommand{\TestRealMOneSig}{$-$0.154}
\newcommand{\TestRealMOneSigHi}{$-$0.074}
\newcommand{\TestRealMOneSigLo}{$-$0.226}
\newcommand{\TestRealMTwoCe}{+0.176}
\newcommand{\TestRealMTwoEe}{+0.441}
\newcommand{\TestRealMTwoNom}{0.989}
\newcommand{\TestRealMTwoSig}{$-$0.265}
\newcommand{\TestRealMTwoSigHi}{$-$0.221}
\newcommand{\TestRealMTwoSigLo}{$-$0.332}
\newcommand{\TestRealMThreeCe}{+0.228}
\newcommand{\TestRealMThreeEe}{+0.420}
\newcommand{\TestRealMThreeNom}{0.992}
\newcommand{\TestRealMThreeSig}{$-$0.191}
\newcommand{\TestRealMThreeSigHi}{$-$0.145}
\newcommand{\TestRealMThreeSigLo}{$-$0.258}
\newcommand{\TestRealMFoureCe}{+0.343}
\newcommand{\TestRealMFoureEe}{+0.441}
\newcommand{\TestRealMFoureNom}{0.997}
\newcommand{\TestRealMFoureSig}{$-$0.099}
\newcommand{\TestRealMFoureSigHi}{$-$0.049}
\newcommand{\TestRealMFoureSigLo}{$-$0.148}
\newcommand{\TestRealMFourjCe}{+0.395}
\newcommand{\TestRealMFourjEe}{+0.438}
\newcommand{\TestRealMFourjNom}{0.994}
\newcommand{\TestRealMFourjSig}{$-$0.043}
\newcommand{\TestRealMFourjSigHi}{0.003}
\newcommand{\TestRealMFourjSigLo}{$-$0.089}
\newcommand{\TestRealDTrainedEEMax}{+0.441}
\newcommand{\TestRealDTrainedEEMin}{+0.312}
\newcommand{\TestRealJudgeCe}{+0.312}
\newcommand{\TestRealJudgeEe}{+0.102}
\newcommand{\TestRealJudgeNom}{0.889}
\newcommand{\TestRealJudgeSig}{+0.210}
\newcommand{\TestRealJudgeSigHi}{0.287}
\newcommand{\TestRealJudgeSigLo}{0.133}
\newcommand{\TestRealNcand}{463}
\newcommand{\TestRealNhard}{324}
\newcommand{\TestRealNvalid}{338}
\newcommand{\TestRealQEight}{125}
\newcommand{\TrAbsentMZero}{0.909}
\newcommand{\TrAbsentMZerolast}{0.912}
\newcommand{\TrAbsentMOne}{0.985}
\newcommand{\TrAbsentMTwo}{0.916}
\newcommand{\TrAbsentMThree}{0.911}
\newcommand{\TrAbsentMFoure}{0.951}
\newcommand{\TrAbsentMFourj}{0.957}
\newcommand{\TrAbsentJudge}{0.990}
\newcommand{\TrRemovalMZero}{0.794}
\newcommand{\TrRemovalMZerolast}{0.888}
\newcommand{\TrRemovalMOne}{0.964}
\newcommand{\TrRemovalMTwo}{0.782}
\newcommand{\TrRemovalMThree}{0.768}
\newcommand{\TrRemovalMFoure}{0.862}
\newcommand{\TrRemovalMFourj}{0.872}
\newcommand{\TrRemovalJudge}{0.937}
\newcommand{\WikiBuilderK}{8}
\newcommand{\WikiBuilderLimit}{1,500}
\newcommand{\WikiBuilderQEight}{185}
\newcommand{\WikiDExtraJudgesN}{4}
\newcommand{\WikiDExtraJudgesPass}{4}
\newcommand{\WikiFictMZeroCe}{+0.401}
\newcommand{\WikiFictMZeroEe}{+0.387}
\newcommand{\WikiFictMZeroNom}{1.000}
\newcommand{\WikiFictMZeroSig}{+0.014}
\newcommand{\WikiFictMZeroSigHi}{0.092}
\newcommand{\WikiFictMZeroSigLo}{$-$0.057}
\newcommand{\WikiFictMZerolastCe}{+0.465}
\newcommand{\WikiFictMZerolastEe}{+0.301}
\newcommand{\WikiFictMZerolastNom}{0.986}
\newcommand{\WikiFictMZerolastSig}{+0.163}
\newcommand{\WikiFictMZerolastSigHi}{0.241}
\newcommand{\WikiFictMZerolastSigLo}{0.085}
\newcommand{\WikiFictMOneCe}{+0.500}
\newcommand{\WikiFictMOneEe}{+0.273}
\newcommand{\WikiFictMOneEeLo}{0.207}
\newcommand{\WikiFictMOneNom}{1.000}
\newcommand{\WikiFictMOneSig}{+0.227}
\newcommand{\WikiFictMOneSigHi}{0.293}
\newcommand{\WikiFictMOneSigLo}{0.161}
\newcommand{\WikiFictMTwoCe}{+0.472}
\newcommand{\WikiFictMTwoEe}{+0.401}
\newcommand{\WikiFictMTwoNom}{0.993}
\newcommand{\WikiFictMTwoSig}{+0.071}
\newcommand{\WikiFictMTwoSigHi}{0.132}
\newcommand{\WikiFictMTwoSigLo}{0.035}
\newcommand{\WikiFictMThreeCe}{+0.479}
\newcommand{\WikiFictMThreeEe}{+0.379}
\newcommand{\WikiFictMThreeNom}{0.993}
\newcommand{\WikiFictMThreeSig}{+0.099}
\newcommand{\WikiFictMThreeSigHi}{0.149}
\newcommand{\WikiFictMThreeSigLo}{0.054}
\newcommand{\WikiFictMFoureCe}{+0.465}
\newcommand{\WikiFictMFoureEe}{+0.465}
\newcommand{\WikiFictMFoureNom}{1.000}
\newcommand{\WikiFictMFoureSig}{+0.000}
\newcommand{\WikiFictMFoureSigHi}{0.043}
\newcommand{\WikiFictMFoureSigLo}{$-$0.043}
\newcommand{\WikiFictMFourjCe}{+0.500}
\newcommand{\WikiFictMFourjEe}{+0.465}
\newcommand{\WikiFictMFourjNom}{1.000}
\newcommand{\WikiFictMFourjSig}{+0.035}
\newcommand{\WikiFictMFourjSigHi}{0.071}
\newcommand{\WikiFictMFourjSigLo}{0.007}
\newcommand{\WikiFictGemmaCe}{+0.362}
\newcommand{\WikiFictGemmaEe}{$-$0.050}
\newcommand{\WikiFictGemmaNom}{0.837}
\newcommand{\WikiFictGemmaSig}{+0.312}
\newcommand{\WikiFictGemmaSigHi}{0.379}
\newcommand{\WikiFictGemmaSigLo}{0.245}
\newcommand{\WikiFictJudgeCe}{+0.429}
\newcommand{\WikiFictJudgeEe}{+0.046}
\newcommand{\WikiFictJudgeNom}{0.922}
\newcommand{\WikiFictJudgeSig}{+0.383}
\newcommand{\WikiFictJudgeSigHi}{0.447}
\newcommand{\WikiFictJudgeSigLo}{0.284}
\newcommand{\WikiFictLlamaCe}{+0.323}
\newcommand{\WikiFictLlamaEe}{+0.053}
\newcommand{\WikiFictLlamaNom}{0.837}
\newcommand{\WikiFictLlamaSig}{+0.270}
\newcommand{\WikiFictLlamaSigHi}{0.355}
\newcommand{\WikiFictLlamaSigLo}{0.163}
\newcommand{\WikiFictNhard}{141}
\newcommand{\WikiFictPhiCe}{+0.379}
\newcommand{\WikiFictPhiEe}{+0.067}
\newcommand{\WikiFictPhiNom}{0.957}
\newcommand{\WikiFictPhiSig}{+0.312}
\newcommand{\WikiFictPhiSigHi}{0.404}
\newcommand{\WikiFictPhiSigLo}{0.199}
\newcommand{\WikiFictQwenCe}{+0.365}
\newcommand{\WikiFictQwenEe}{$-$0.004}
\newcommand{\WikiFictQwenNom}{0.894}
\newcommand{\WikiFictQwenSig}{+0.362}
\newcommand{\WikiFictQwenSigHi}{0.397}
\newcommand{\WikiFictQwenSigLo}{0.248}
\newcommand{\WikiRealMZeroCe}{+0.303}
\newcommand{\WikiRealMZeroEe}{+0.381}
\newcommand{\WikiRealMZeroNom}{0.993}
\newcommand{\WikiRealMZeroSig}{$-$0.078}
\newcommand{\WikiRealMZeroSigHi}{$-$0.017}
\newcommand{\WikiRealMZeroSigLo}{$-$0.139}
\newcommand{\WikiRealMZerolastCe}{+0.385}
\newcommand{\WikiRealMZerolastEe}{+0.225}
\newcommand{\WikiRealMZerolastNom}{0.976}
\newcommand{\WikiRealMZerolastSig}{+0.159}
\newcommand{\WikiRealMZerolastSigHi}{0.232}
\newcommand{\WikiRealMZerolastSigLo}{0.093}
\newcommand{\WikiRealMOneCe}{+0.425}
\newcommand{\WikiRealMOneEe}{+0.314}
\newcommand{\WikiRealMOneNom}{0.990}
\newcommand{\WikiRealMOneSig}{+0.112}
\newcommand{\WikiRealMOneSigHi}{0.169}
\newcommand{\WikiRealMOneSigLo}{0.063}
\newcommand{\WikiRealMTwoCe}{+0.446}
\newcommand{\WikiRealMTwoEe}{+0.392}
\newcommand{\WikiRealMTwoNom}{1.000}
\newcommand{\WikiRealMTwoSig}{+0.054}
\newcommand{\WikiRealMTwoSigHi}{0.086}
\newcommand{\WikiRealMTwoSigLo}{0.005}
\newcommand{\WikiRealMThreeCe}{+0.449}
\newcommand{\WikiRealMThreeEe}{+0.371}
\newcommand{\WikiRealMThreeNom}{1.000}
\newcommand{\WikiRealMThreeSig}{+0.078}
\newcommand{\WikiRealMThreeSigHi}{0.115}
\newcommand{\WikiRealMThreeSigLo}{0.030}
\newcommand{\WikiRealMFoureCe}{+0.453}
\newcommand{\WikiRealMFoureEe}{+0.419}
\newcommand{\WikiRealMFoureNom}{0.993}
\newcommand{\WikiRealMFoureSig}{+0.034}
\newcommand{\WikiRealMFoureSigHi}{0.071}
\newcommand{\WikiRealMFoureSigLo}{$-$0.003}
\newcommand{\WikiRealMFourjCe}{+0.466}
\newcommand{\WikiRealMFourjEe}{+0.446}
\newcommand{\WikiRealMFourjNom}{0.997}
\newcommand{\WikiRealMFourjSig}{+0.020}
\newcommand{\WikiRealMFourjSigHi}{0.058}
\newcommand{\WikiRealMFourjSigLo}{$-$0.014}
\newcommand{\WikiRealDJudgeRankNominal}{8}
\newcommand{\WikiRealDJudgeRankSigma}{1}
\newcommand{\WikiRealDNRankedSystems}{8}
\newcommand{\WikiRealDNTrained}{6}
\newcommand{\WikiRealDNominalMax}{1.000}
\newcommand{\WikiRealDNominalMin}{0.905}
\newcommand{\WikiRealDTrainedCEMax}{+0.466}
\newcommand{\WikiRealDTrainedCEMin}{+0.303}
\newcommand{\WikiRealDTrainedSigmaPosCount}{5}
\newcommand{\WikiRealGemmaCe}{+0.254}
\newcommand{\WikiRealGemmaEe}{$-$0.005}
\newcommand{\WikiRealGemmaNom}{0.780}
\newcommand{\WikiRealGemmaSig}{+0.249}
\newcommand{\WikiRealGemmaSigHi}{0.286}
\newcommand{\WikiRealGemmaSigLo}{0.185}
\newcommand{\WikiRealJudgeCe}{+0.327}
\newcommand{\WikiRealJudgeEe}{+0.164}
\newcommand{\WikiRealJudgeNom}{0.905}
\newcommand{\WikiRealJudgeSig}{+0.163}
\newcommand{\WikiRealJudgeSigHi}{0.234}
\newcommand{\WikiRealJudgeSigLo}{0.085}
\newcommand{\WikiRealLlamaCe}{+0.232}
\newcommand{\WikiRealLlamaEe}{+0.093}
\newcommand{\WikiRealLlamaNom}{0.797}
\newcommand{\WikiRealLlamaSig}{+0.139}
\newcommand{\WikiRealLlamaSigHi}{0.224}
\newcommand{\WikiRealLlamaSigLo}{0.051}
\newcommand{\WikiRealNhard}{295}
\newcommand{\WikiRealPhiCe}{+0.276}
\newcommand{\WikiRealPhiEe}{+0.120}
\newcommand{\WikiRealPhiNom}{0.885}
\newcommand{\WikiRealPhiSig}{+0.156}
\newcommand{\WikiRealPhiSigHi}{0.237}
\newcommand{\WikiRealPhiSigLo}{0.078}
\newcommand{\WikiRealQwenCe}{+0.324}
\newcommand{\WikiRealQwenEe}{+0.134}
\newcommand{\WikiRealQwenNom}{0.885}
\newcommand{\WikiRealQwenSig}{+0.190}
\newcommand{\WikiRealQwenSigHi}{0.265}
\newcommand{\WikiRealQwenSigLo}{0.112}

\date{2026-11-08}   

\title{ChainCheck: Near-Perfect Evidence-Sufficiency Scores\\ Can Conceal Sensitivity to Construction Artifacts\\[4pt]
\large Paired Conditional Contrasts that Expose an Evaluation Confound}
\author{Hyojung Han$^{1,\dagger}$ \quad Ryangkyung Kang$^{1}$ \quad Youngeun Kim$^{2}$
\quad Seung-Hun Jeon$^{1}$\\[4pt]
\small $^{1}$ThakiCloud \qquad $^{2}$Korea University\\[2pt]
\small $^{\dagger}$Corresponding author: \texttt{hyojung.han@thakicloud.com}}

\begin{document}
\maketitle

\begin{abstract}
Evidence-sufficiency benchmarks ask whether a model can tell a complete passage set from one whose
reasoning chain is broken. When broken sets are made by editing text, near-perfect sufficiency
scores---and even positive crossed-control scores---need not demonstrate multi-hop reasoning. We present
ChainCheck, an evaluation-validity audit that pairs validator-certified counterfactual cells with
artifact-only baselines. In an exploratory crossed control on \CxNE{} HotpotQA pairs, a language-model
scorer that never reads the chain reaches a nominal AUC of \CxAbtwoNom{}, and a one-string match reaches
the maximum of the crossed chain-sensitivity statistic CS4 (\CxAbthreeCs{}), while the \CxNCs{} evaluated
systems score from \CxCsMin{} to \CxCsMax{}; the two statistics are different metrics. We therefore read
positive CS4 as operational bridge-consistency sensitivity, not as semantic multi-hop reasoning. Every
system is also sensitive to one extra edit, and the edits are detectable according to model-based
naturalness ratings. On an independent HotpotQA set of \HpN{} pairs, patterns pre-specified on MuSiQue
and 2WikiMultiHopQA did not replicate: \HpHOnePass{} of \HpHOneN{} trained readouts were edit-dominated,
and the reference judge's selectivity interval [\HpJudgeSigLo, \HpJudgeSigHi] includes zero. Judge scores
also depended on batch layout. The MuSiQue and 2Wiki records are public; the HotpotQA records and the
fine-tuned checkpoints are not, so external reproduction is partial.
\end{abstract}

\section{Introduction}

How do we know that an evidence-sufficiency model detects broken reasoning chains rather than artifacts
left behind when the benchmark was built? A sufficiency benchmark needs negatives: passage sets that
look like the positives but lack a complete evidence chain. Every way of making them leaves a trace.
Deleting a supporting passage lowers the query relevance of the set, so a surface classifier recovers
the label. Substituting the bridge entity in a later hop keeps query overlap fixed, but it places an
unexpected entity in a fluent passage, and a large language model can notice that the text was edited
without following the chain at all. The ordinary score, a pair AUC of positive over negative, mixes these
behaviours. Once several systems sit near the ceiling, it also leaves little headroom to order them.

Shortcut learning is well documented \cite{geirhos2020shortcut}, and counterfactual evaluation of
multi-hop reasoning is not new. ACCORD controls reasoning depth with multi-hop counterfactuals
\cite{roewer2025accord}; CRiT-QA and CofCA use counterfactual entities and chains to remove parametric
knowledge and shallow shortcuts \cite{yun2026critqa,wu2024cofca}; SURE-RAG validates a sufficiency
verifier under an artifact-aware protocol \cite{qiu2026surerag}; and Pair-ID audits how a generated answer
responds to paired evidence interventions \cite{du2026pairid}. These works use counterfactual edits as a
tool. ChainCheck asks whether the counterfactual edit itself becomes a new shortcut. Rather than
proposing a harder benchmark, it reports, next to the ordinary score, a chain effect, an edit effect and
chain selectivity, each a paired conditional contrast on pairs whose labels follow mechanically from validated
provenance-preserving interventions. We position it as a protocol that exposes an evaluation confound: a
positive selectivity says that a system's response on these records is not explained by the edit trace
alone, not that the system understands the chain.

The protocol grew out of a failure. We had pre-specified a comparison between a structured evidence
readout and a single-layer probe on a provenance-built benchmark. The headline gate failed because the
probe was already at the ceiling, and a matched-adaptation comparison against a one-token judge ended in a
tie (Section~\ref{sec:ceiling}). Controls added before the test read showed that much of the probe's
score was edit detection. ChainCheck formalises those controls into a protocol.

\begin{figure}[!htbp]
\centering
\begin{minipage}[t]{\dimexpr\textwidth/2-1em\relax}\centering
\includegraphics[width=\linewidth]{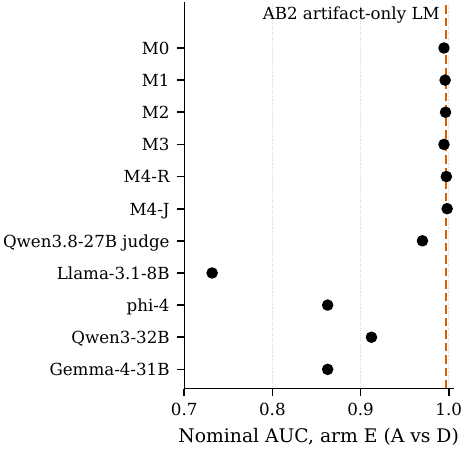}\\[-2pt]
{\small (a) Nominal AUC}
\end{minipage}\hfill
\begin{minipage}[t]{\dimexpr\textwidth/2-1em\relax}\centering
\includegraphics[width=\linewidth]{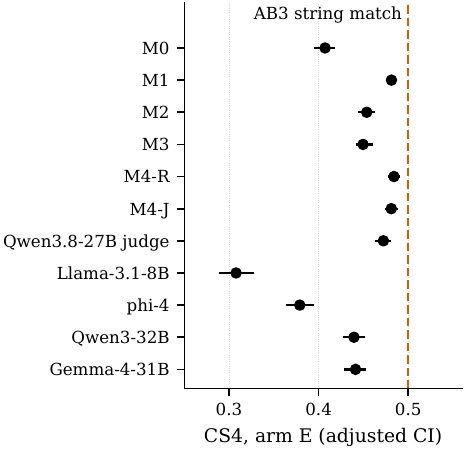}\\[-2pt]
{\small (b) CS4}
\end{minipage}
\caption{Artifact-only baselines against the evaluated systems on the crossed HotpotQA arm E
(\CxNE{} pairs, exploratory). (a) Nominal pair AUC of the original contrast (A over D) for the \CxFamM{}
systems; the dashed line is AB2, a per-passage language-model log-probability scorer that never sees the
chain (\CxAbtwoNom{}). (b) CS4, the crossed chain-sensitivity statistic of Section~\ref{sec:crossed}, with
Bonferroni-adjusted document-group bootstrap intervals; the dashed line is AB3, which checks whether one
bridge string equals one title and reaches the maximum \CxAbthreeCs{}. The panels use different metrics on
separate axes and are not compared with each other. Values are in Table~\ref{tab:crossed}.}
\label{fig:headline}
\end{figure}

\paragraph{Contributions.}
\begin{enumerate}
\item \emph{A validity problem for sufficiency scores.} Near-ceiling nominal scores and positive
crossed-control scores can both be reached without reading the chain. On the crossed HotpotQA arm the
evaluated systems' nominal AUC lies between \CxNomMin{} and \CxNomMax{}, an artifact-only language-model
scorer reaches \CxAbtwoNom{}, and a string-match reference reaches the CS4 maximum of \CxAbthreeCs{}
(Figure~\ref{fig:headline}, Section~\ref{sec:crossed}). We separate three claims: sensitivity to bridge
string consistency (supported), sensitivity to the edit trace (supported) and semantic multi-hop
reasoning (not established by any measurement here).
\item \emph{An audit protocol, with its boundary.} ChainCheck admits a pair only when a deterministic
validator (checks Q1 to Q9) certifies every counterfactual cell, reports a chain effect, an edit effect
and their difference as paired conditional contrasts, and adds a crossed control with artifact-only
baselines (Sections~\ref{sec:protocol} and~\ref{sec:crossed}). Its boundary is stated with it: a broken
chain without any edit cannot be constructed, so the contrasts are conditional, not an additive
decomposition, and CS4 measures operational chain sensitivity in the sense defined in
Section~\ref{sec:crossed}.
\item \emph{Limits on generalization and reproduction.} Patterns pre-specified on MuSiQue
(Section~\ref{sec:confirm}) and tested on 2WikiMultiHopQA (Section~\ref{sec:wiki}) do not replicate on an
independent HotpotQA set whose clustered intervals are informative (Section~\ref{sec:hotpotconf}); record
dependence limits the earlier sets (Section~\ref{sec:posthoc}); judge scores depend on batch layout
(Section~\ref{sec:batch}); and the data and fine-tuned checkpoints are private, which bounds what outside
researchers can reproduce (Section~\ref{sec:avail}).
\end{enumerate}

\paragraph{Status of the evidence.} Section~\ref{sec:explore} is exploratory: selectivity and its
criterion were defined after the single read of the MuSiQue test split. Before scoring any confirmation
pair we committed three hypotheses, their thresholds and the system list to a time-stamped internal decision record (not public)
(Section~\ref{sec:confirm}), and before scoring any 2Wiki pair we fixed the replication patterns and the
only permitted builder adaptations (Section~\ref{sec:wiki}). The headline wording of this paper was
narrowed after the 2Wiki result; no criterion, number or system choice changed. Section~\ref{sec:posthoc}
was added after every pre-specified result had been read; it changes no pre-specified verdict, and its
informativeness rule was itself chosen post hoc. Section~\ref{sec:hotpot} was run after all of these
results had been read; the tier of each of its analyses is stated there (Table~\ref{tab:tiers}).

\section{Related work}

\paragraph{Shortcuts and annotation artifacts.} Models often reach high benchmark scores through cues that
the benchmark designers did not intend \cite{geirhos2020shortcut}. Hypothesis-only baselines in natural
language inference reach well above chance \cite{gururangan2018annotation,poliak2018hypothesis}, models
trained on that task adopt syntactic heuristics that fail on targeted evaluation sets \cite{mccoy2019right},
and reading-comprehension models answer many questions from the question or the passage alone
\cite{kaushik2018reading}. ChainCheck addresses one artifact of this kind that is specific to sufficiency
benchmarks: the trace that constructing a negative leaves in the text.

\paragraph{Shortcuts in multi-hop question answering.} HotpotQA introduced explainable multi-hop questions
with supporting facts \cite{yang2018hotpotqa}, and several analyses then showed that many of its questions
can be answered with a single hop \cite{min2019compositional,chen2019understanding}. Adversarial
distractors \cite{jiang2019avoiding} and transformations that expose disconnected reasoning
\cite{trivedi2020dire} were proposed to close these shortcuts, and models that answer a multi-hop question
correctly often fail its single-hop sub-questions \cite{tang2020multihop}. Language models show a
compositionality gap between answering sub-questions and composing them \cite{press2023measuring}.
MuSiQue composes multi-hop questions from single-hop ones so that each hop is needed and records per-hop
supporting paragraphs \cite{trivedi2022musique}; 2WikiMultiHopQA provides evidence triples for each
reasoning path \cite{ho2020constructing}. We use that provenance as the only source of labels.

\paragraph{Counterfactual and minimal-edit evaluation.} Contrast sets \cite{gardner2020contrast} and
counterfactually augmented data \cite{kaushik2020learning} use minimal edits to probe or train decision
boundaries. For multi-hop reasoning, ACCORD uses controlled multi-hop anti-factuals to separate commonsense
grounding from reasoning and to vary reasoning complexity \cite{roewer2025accord}; ACCORD controls what
reasoning is required, whereas ChainCheck controls what construction artifact accompanies the reasoning
intervention. CofCA builds step-wise counterfactual questions \cite{wu2024cofca}, and CRiT-QA replaces
factual chains with counterfactual entities and adds distractor chains that diverge at different hops
\cite{yun2026critqa}. CRiT-QA uses counterfactual edits to remove shortcuts; ChainCheck measures whether the
counterfactual edit itself becomes a shortcut. It adds a cell in which the edit is present but the label
does not change, and reports the response to the chain and the response to the edit as separate
quantities.

\paragraph{Knowledge conflicts.} When context contradicts what a model memorised, models may follow the
context or their parametric knowledge \cite{longpre2021entity,xie2024adaptive}. Our consistent bridge swap
with a real entity creates exactly such a conflict in the earlier hop, which is why the protocol includes a
synthetic-entity control.

\paragraph{Evidence sufficiency in retrieval-augmented generation.} SQuAD 2.0 made unanswerable questions
a reading-comprehension construct \cite{rajpurkar2018know}, and sufficient context has been proposed as a
lens on retrieval-augmented generation \cite{joren2025sufficient}. Benchmarks and analyses of
retrieval-augmented generation test robustness to noisy, irrelevant or missing evidence
\cite{chen2024benchmarking,cuconasu2024power,yoran2024making}, SURE-RAG treats evidence sufficiency as a set-level property and validates a
proposed sufficiency verifier with counterfactual swaps, shortcut baselines and no-oracle checks
\cite{qiu2026surerag}. ChainCheck asks a different question: whether the counterfactual construction itself
creates a signal that can masquerade as sufficiency detection. It therefore adds an intact-chain edit
control and measures chain and edit sensitivity separately for any evaluated scorer, rather than proposing
a scorer. Pair-ID holds a query, retrieval state and reader fixed and crosses support addition with
nonsupport deletion to audit how a failed answer responds to evidence repair \cite{du2026pairid}. Its
crossed design resembles ours, but its object is the generated response; ChainCheck's object is the
evaluation itself, asking how much of a sufficiency score responds to the benchmark's construction edit
rather than to the broken chain. Pair-ID is a response audit; ChainCheck is an evaluation-confound audit.

\paragraph{Probes and judges.} Probing classifiers can report high accuracy for reasons unrelated to the
property of interest, which control tasks were designed to expose \cite{hewitt2019designing}; our edit-only
cell plays a similar role for sufficiency readouts. Language models are widely used as zero-shot judges
\cite{zheng2023judging}; here they are evaluated systems, not labelers. LLM safety evaluators shift their
preferences with superficial artifacts such as apologetic phrasing, and agreement with human labels does
not imply robustness to such artifacts \cite{chen2025safer}. ChainCheck makes the same point for
evidence-sufficiency scorers, with a construction edit in place of a stylistic one.

\section{The ChainCheck protocol}\label{sec:protocol}

\subsection{Counterfactual cells}
Cells are indexed by chain status (alive, broken) and edit trace (absent, present). Three of the four
cells are constructed; the design is therefore not a full factorial.

\begin{center}\small
\begin{tabular}{lcc}
\toprule
 & No edit trace & Edit trace \\
\midrule
Chain alive & A: original passage set & B: consistent bridge swap \\
Chain broken & C: not constructed & D: bridge substitution in the later hop \\
\bottomrule
\end{tabular}
\end{center}

D replaces the bridge entity $X$ with a different entity $Y$ in the later-hop supporting passage only, so
the earlier hop yields $X$, the later hop speaks of $Y$, and the chain is cut. B replaces $X$ with $Y$ in
both the earlier and the later supporting passages, so the chain is intact but carries an edit trace. B
and D share the later passage exactly and differ only in the earlier supporting passage. C, a broken chain
without any edit, cannot be built without an intervention and is not used. Removing a hop is the nearest
candidate, but it carries a different artifact (lower query relevance), so hop removal appears only as a
diagnostic. Because C is missing, ChainCheck cannot estimate an interaction between chain status and edit
trace, and the chain effect is measured in the presence of an edit.

\subsection{Metrics}
The primitive statistic is a within-record concordance probability over the $n$ accepted records,
$\mathrm{AUC}(X>Y)=n^{-1}\sum_i\bigl[\mathbf{1}(s_{X_i}>s_{Y_i})+\tfrac{1}{2}\,\mathbf{1}(s_{X_i}=s_{Y_i})\bigr]$,
which we call a pair AUC; it is invariant to score scale and calibration. CE, EE and $\Sigma$ are
contrasts derived from it, not AUCs themselves. They are paired conditional contrasts, not an additive
decomposition: AUC is nonlinear and C is absent, so the nominal score is not
the sum of the chain and edit effects. Each contrast is identified only conditionally: CE is the response
to a broken chain given that an edit is present, and EE is the response to an edit given that the chain is
alive. No interaction between chain status and edit trace is estimable from three cells, and we do not
read CE and EE as components of the nominal score.
\begin{align*}
\text{Nominal} &= \mathrm{AUC}(A > D), &
\text{CE} &= \mathrm{AUC}(B > D) - 0.5,\\
\text{EE} &= \mathrm{AUC}(A > B) - 0.5, &
\Sigma &= \text{CE} - |\text{EE}|.
\end{align*}
The chain effect CE is the B-versus-D response to breaking the chain under a shared later-passage edit; the edit
effect EE is the response to an edit that leaves the chain alive; chain selectivity $\Sigma$ is positive
when chain sensitivity exceeds edit sensitivity. A system is chain-selective on a slice when the lower
bound of the 95\% pair-bootstrap interval of $\Sigma$ is above zero. Intervals are 2{,}000-resample
percentile bootstraps over accepted records. For trained systems, displayed point estimates are medians of
three seed-level metrics, whereas the bootstrap resamples per-record outcomes averaged over the seeds; the
intervals therefore target a seed-averaged statistic, are not uncertainty intervals for the displayed
medians, and do not include training-seed variation. The chain-selectivity criterion and H1 were applied
to this seed-averaged bootstrap. B edits two passages and D edits one,
so for a scorer that responds only to the number of edited passages this asymmetry would push CE
downward. No general bias direction follows for other scorers, because B and D may also differ in edit
coherence, naturalness, or chain-correlated surface cues. The same mismatch cuts the other way for EE: A
versus B compares an unedited set with a two-passage edit, so EE may exceed the edit effect of a
single-passage edit, and $|\text{EE}|$ enters $\Sigma$ at that larger value. Neither contrast is
edit-matched, so CE is not a chain-only effect. A follow-up edit-matched control confirms the predicted
direction: matching raises CE for every system on MuSiQue, most for the trained readouts
(Section~\ref{sec:open}).

\subsection{Deterministic validator}
A pair enters the counts only if nine checks pass: Q1 the query is identical across cells; Q2 passage
positions are identical; Q3 untouched passages are byte-identical; Q4 the later passage is identical in B
and D; Q5 the earlier-passage edit in B is exactly the $X$ to $Y$ replacement; Q6 $X$ and $Y$ are
alias-disjoint; Q7 the answer-span count is identical; Q8 each passage's query-lexical signature is
identical across cells; Q9 the chain graph is alive in A and B and broken in D. Labels come only from
dataset provenance and these checks. Our claim is restricted to programmatically verifiable
provenance-preserving interventions: the benchmark labels follow mechanically from validated
interventions, and human audit is not part of the protocol. The checks are string- and graph-level: they
do not certify that an edited passage reads naturally or that the substituted entity is plausible in
context, so any such cue counts as part of the edit trace rather than being removed.

\subsection{Controls}
(1) Closed-book separation: questions the reference backbone answers without passages, under a lenient
string-containment rule, are excluded. This reduces but does not eliminate parametric bypass, and it
does not establish closed-book hardness for the additional judge families. (2) Edit-only control: cell B, measured by EE. (3)
Conditional chain contrast: B against D, measured by CE. (4) Synthetic entities: $Y$ is a deterministic
fictional name of matched token count and capitalisation, which reduces bridge-specific world-knowledge conflict in B
and D. (5) Permutation invariance: passage order is shuffled and features are re-extracted from the
language model, not just re-pooled. A diagnostic, construction transfer, scores systems trained on
substitution against negatives built by hop removal or full removal.

\section{Construction}\label{sec:construction}

\paragraph{MuSiQue.} We group examples by a hash of their sorted supporting document titles and split by
group. The original MuSiQue split is not document-disjoint: of \GrpTrainDocs{} train and \GrpValDocs{}
validation documents, \GrpSharedDocs{} are shared, and only \GrpValDisjoint{} of \GrpValTotal{}
validation examples (\GrpDisjointPct\%) are fully disjoint; we keep only those for test. An earlier build
took the group identifier from the prefix of the example id, which encodes the hop count; that produced
\GrpGroupsBefore{} groups and silently turned group-aware cross-validation into ordinary k-fold. The
corrected definition gives \GrpGroupsAfter{} groups, at most \GrpMaxGroup{} examples per group and
\GrpSingletonPct\% singletons, and re-measuring the surface gate moved it from \GrpGZeroBeforeFix{} to
\SurfSubstTest{}.

Table~\ref{tab:surface} shows why the negatives are built by substitution: a surface-only classifier
separates deletion-based negatives from the original set but not substitution-based ones. On all
\BenchTest{} test pairs the deltas of maximum and mean query Jaccard, answer presence and answer count are
exactly zero; the BM25 maximum changes for \SurfBmTwoFiveNonzeroPct\% of pairs. A legacy eight-invariant
builder gate, applied before the Q1--Q9 validator, decides which substitution candidates enter the
benchmark; \CvTrainClean{} of \CvTrainChecked{}
train candidates (\CvTrainPct\%) and \CvValClean{} of \CvValChecked{} validation candidates (\CvValPct\%)
pass. It catches two corruptions that surface audits cannot see: a bridge that is part of the answer
string, so that substitution also edits the answer (\CvTrainVSix{} train candidates), and an intermediate
passage that mentions the bridge and rebuilds the chain (\CvTrainVThree{} train and \CvValVThree{}
validation candidates). The primary arm has \BenchTrain{} train, \BenchDev{} dev, \BenchCalib{} calibration
and \BenchTest{} test pairs.

\begin{table}[t]
\centering\small
\begin{tabular}{lcc}
\toprule
Construction of the negative (MuSiQue test) & Surface-only paired AUC & Role \\
\midrule
ABSENT, all support removed & \SurfAbsentAuc & diagnostic \\
PARTIAL, hop removal & \SurfRemovalAuc & diagnostic \\
PARTIAL, bridge substitution & \SurfSubstTest & primary \\
\bottomrule
\end{tabular}
\caption{A surface-only classifier (BM25, Jaccard, length, near-duplicate features) under group-aware cross-validation. Deletion-based negatives are separable from the surface alone; bridge substitution is below the pre-specified gate of \GateGZeroMax.}
\label{tab:surface}
\end{table}

On the test split, \TestRealNvalid{} of \TestRealNcand{} real-entity candidates and \TestFictNvalid{} of
\TestFictNcand{} synthetic-entity candidates (the pairs admitted by the builder gate) pass Q1 to Q9, of which \TestRealNhard{} and \TestFictNhard{}
are closed-book-hard. Every rejection was Q8 (\TestRealQEight{} real and \TestFictQEight{} synthetic): the
earlier-passage edit in B had changed that passage's query-lexical signature. Kept, those pairs would have
turned part of the A-versus-B difference into a query-relevance artifact, the confound the protocol exists
to remove. Only the validator caught it.

\paragraph{MuSiQue confirmation.} The confirmation set takes MuSiQue validation examples that the test
split excluded for sharing documents with training, built by the same builder, controls and validator
(\MuRealNhard{} real and \MuFictNhard{} synthetic closed-book-hard pairs). Because these examples share
documents with the training split, confirmation rests on the untrained judge and on contrasts, not on
absolute scores of trained systems; the hypothesis about trained systems (H1 below) is a within-distribution
check rather than an independent confirmation.

\paragraph{2WikiMultiHopQA.} The only permitted changes were to the builder, and they were fixed before
any score: the bridge is the object of the first evidence triple and the subject of the second; only
compositional and inference questions are used; each set has \WikiBuilderK{} passages because 2Wiki gives
ten per question; and at most \WikiBuilderLimit{} examples are drawn in a fixed seeded order. The
validator rejected \WikiBuilderQEight{} candidates for Q8, leaving \WikiRealNhard{} real and
\WikiFictNhard{} synthetic closed-book-hard pairs. Passages are rendered as a title followed by the body,
and in every admitted 2Wiki example the later-hop passage title is the bridge entity, so the bridge
substitution also rewrites that title prefix in B and D; MuSiQue passages carry titles too, but the
later-hop title is not in general the bridge.

\section{Systems evaluated}\label{sec:systems}

The eight primary systems use the same Qwen3.8-27B base checkpoint, with the passages packed into one
prompt that ends in a decision token. M0 to M3 and the zero-shot judge keep the backbone frozen; M4-R and
M4-J add matched rank-8 LoRA adapters. The systems are subjects of the evaluation; none is proposed here.
\textbf{M0} is a logistic probe on the decision token of the single layer that is best on dev (layer
\MZeroBestLayer, dev pair AUC \MZeroBestAuc); \textbf{M0-last} uses the last layer. \textbf{M1} mixes
sixteen layers with a softmax gate. \textbf{M2} pools per-passage representations with attention and a
learned query, without cross-passage interaction. \textbf{M3} lets the mixed decision token attend over
the passage set without positional encoding. \textbf{M4-R} and \textbf{M4-J} spend the same rank-8 LoRA
budget \cite{hu2022lora} (\MFourLoraParams{} parameters on the full-attention layers) to feed a readout head or to score a
one-token answer. The \textbf{zero-shot judge} is the backbone's one-token yes/no probability, untrained.
M0 to M3 are trained with binary cross-entropy and a pair hinge on primary substitution pairs only (A
against D), early-stopped on dev, and reported as the median of three seeds. The M4 adapters train on
\MFourTrainPairs{} pairs. For 2Wiki, every trained system is the MuSiQue-trained model; nothing is
retrained or recalibrated.

\paragraph{Additional judges.} Separately from the eight primary systems, we score four zero-shot judges
from other model families on both evaluation sets, with the same prompt and score definition:
Llama-3.1-8B-Instruct, phi-4, Qwen3-32B and Gemma-4-31B-it. They are not part of the primary ranking.

\section{Results}\label{sec:results}

\subsection{The nominal ceiling}\label{sec:ceiling}
On the \FinNHard{} closed-book-hard MuSiQue test pairs (the backbone answered \FinExposedPct\% of the
questions closed-book), the nominal pair AUC is \FinNomMZero{} for M0, \FinNomMThree{} for M3,
\FinNomMFoure{} for both M4 systems and \FinNomJudge{} for the zero-shot judge. The pre-specified gate
requiring M3 to beat M0 by at least \FinGbMargin{} failed (gain \FinGbDelta{} [\FinGbLo, \FinGbHi]), and
the matched-adaptation comparison was a tie (difference \FinGcDelta{} [\FinGcLo, \FinGcHi]). These ceiling-level
values leave limited headroom, and we do not report inferential comparisons of the nominal ranks. We report these outcomes as the observation that
motivated the protocol, not as contributions. This section reports the original pre-specified A-versus-D
test on \FinNHard{} closed-book-hard pairs. ChainCheck additionally requires a valid B cell and checks Q1 to
Q9, which leaves \TestRealNhard{} real-entity pairs for the sections that follow.

\subsection{Exploratory contrasts on the MuSiQue test split}\label{sec:explore}
\begin{figure}[!htbp]
\centering
\includegraphics[width=\textwidth]{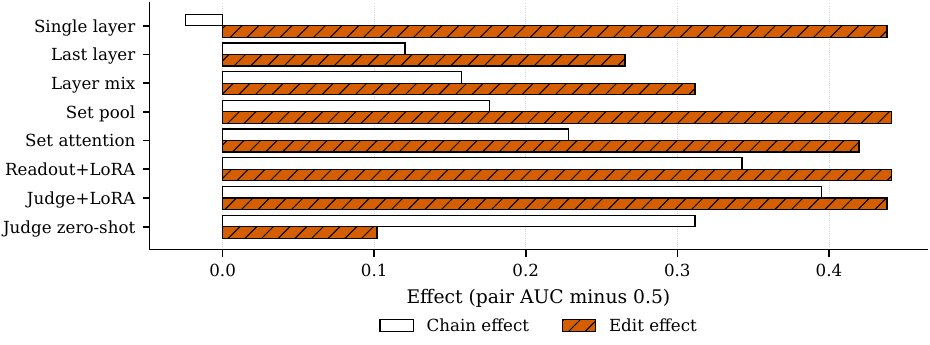}\\[2pt]
\small (a) MuSiQue test (exploratory)\\[4pt]
\includegraphics[width=\textwidth]{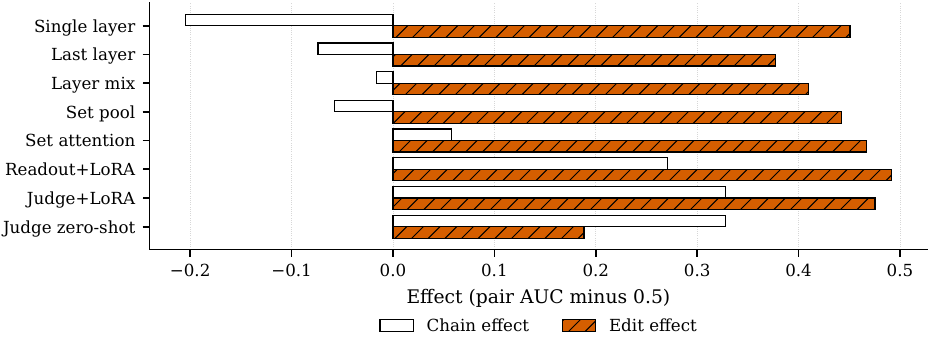}\\[2pt]
\small (b) MuSiQue confirmation\\[4pt]
\includegraphics[width=\textwidth]{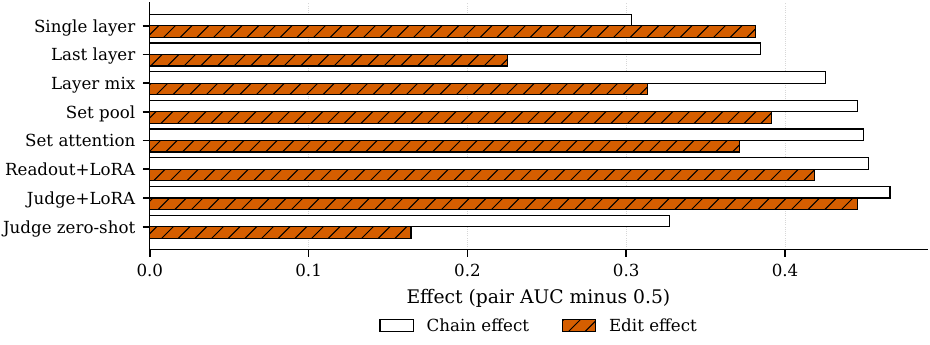}\\[2pt]
\small (c) 2WikiMultiHopQA replication
\caption{Chain effect (open bars, AUC(B$>$D)$-0.5$) and edit effect (hatched bars, AUC(A$>$B)$-0.5$) per
system, real bridge entities. Trained systems (M0 to M4) show large edit effects in every setting; their
chain effects are smaller on MuSiQue and large on 2Wiki. The two contrasts vary separately in these data.}
\label{fig:decomp}
\end{figure}

Table~\ref{tab:real} and Figure~\ref{fig:decomp} give the contrasts for every system on the
\TestRealNhard{} real-entity test pairs (\TestFictNhard{} synthetic). On the test split
every system trained on A versus D treats the intact but edited chain B almost as a negative: edit effects
lie between \TestRealDTrainedEEMin{} and \TestRealDTrainedEEMax{}. The single-layer probe shows no chain
sensitivity at all (chain effect \TestRealMZeroCe{} [\TestRealMZeroCeLo, \TestRealMZeroCeHi]), structured
readouts and adaptation raise the chain effect, and the edit effect stays larger, so every trained
system has negative selectivity. The zero-shot judge is the only chain-selective system, \TestRealJudgeSig{}
[\TestRealJudgeSigLo, \TestRealJudgeSigHi], while having the lowest nominal score, \TestRealJudgeNom{}. With
synthetic bridge names (Table~\ref{tab:fict}) the judge's edit effect falls to \TestFictJudgeEe{}
[\TestFictJudgeEeLo, \TestFictJudgeEeHi], which is consistent with a contribution from bridge-specific world-knowledge conflict to its
residual edit sensitivity, whereas the trained systems keep their edit effect. Selectivity and the
chain-selectivity criterion were defined after this read, so this section motivates the hypotheses
tested next and does not confirm them.

\subsection{Sealed confirmation on MuSiQue}\label{sec:confirm}
Three hypotheses were frozen before any confirmation pair was scored. H1: every trained system has an
edit effect whose lower bound exceeds 0.25 and a selectivity whose upper bound is below zero. H2: the
reference zero-shot judge has a selectivity lower bound above zero on both the real and the synthetic
slice. H3: the judge's edit effect is larger on real than on synthetic entities. H1 and H2 met their
pre-specified criteria. H3 had the pre-specified direction, but its independent-slice bootstrap ignores shared
source questions and documents; we report H3 descriptively and do not assign it a confirmatory
pass/fail interpretation. H1
concerns trained systems scored on documents shared with their training split; it is a
within-distribution check, not an independent confirmation.

On real entities (\MuRealNhard{} pairs; \MuFictNhard{} synthetic), the lower bounds of the trained
systems' edit effects lie between
\MuRealDTrainedEELoMin{} and \MuRealDTrainedEELoMax{}, and every trained selectivity interval lies below
zero. The reference judge's selectivity is \MuRealJudgeSig{} [\MuRealJudgeSigLo, \MuRealJudgeSigHi] on real
and \MuFictJudgeSig{} [\MuFictJudgeSigLo, \MuFictJudgeSigHi] on synthetic entities; the margin on real
entities is thin. Its edit effect is \MuHThree{} [\MuHThreeLo, \MuHThreeHi] lower on the synthetic slice than on the
real slice. The interval is the difference of independent bootstraps of two separately validated slices
whose membership differs (\MuRealNhard{} and \MuFictNhard{} pairs), so the difference is consistent with,
but not identified as, a contribution from bridge-specific world-knowledge conflict to its residual edit
sensitivity.

One exploratory observation does not carry over. On the test split the judge had the lowest nominal
score; on the confirmation set it ranks \MuRealDJudgeRankNominal{} of \MuRealDNRankedSystems{} by nominal
score and \MuRealDJudgeRankSigma{} by selectivity. What holds is that nominal scores lie between
\MuRealDNominalMin{} and \MuRealDNominalMax{} on real entities and leave little headroom, while
the selectivity point estimates order them differently, with overlapping intervals.

\paragraph{Other judge families.} Following a rule fixed before their scores were seen, every
additional judge is reported on both datasets (Tables~\ref{tab:judges} and~\ref{tab:judges_fict}). On the
MuSiQue confirmation set, \MuDExtraJudgesPass{} of the \MuDExtraJudgesN{} additional judges meet the H2
criterion on both slices: Gemma-4-31B-it is chain-selective, while phi-4 is edit-dominated on real
entities, with selectivity \MuRealPhiSig{} [\MuRealPhiSigLo, \MuRealPhiSigHi]. On 2Wiki all
\WikiDExtraJudgesPass{} of \WikiDExtraJudgesN{} meet it, phi-4 included (\WikiRealPhiSig{}
[\WikiRealPhiSigLo, \WikiRealPhiSigHi] on real entities). The pre-specified reference judge replicates,
while selectivity varies across judge families and datasets. Being untrained on the nominal contrast is
therefore not by itself what makes a system chain-selective, which is what one expects if the two contrasts are
separable.

\subsection{External replication on 2WikiMultiHopQA}\label{sec:wiki}
The replication was run because H1 and H2 passed, as pre-specified, and was required to reproduce four
patterns rather than absolute numbers. Three replicate on the \WikiRealNhard{} real-entity pairs
(\WikiFictNhard{} synthetic). Nominal scores lie between \WikiRealDNominalMin{}
and \WikiRealDNominalMax{}, and the judge ranks \WikiRealDJudgeRankNominal{} of \WikiRealDNRankedSystems{}
(last) by nominal score on both slices. Its selectivity is positive, \WikiRealJudgeSig{} [\WikiRealJudgeSigLo,
\WikiRealJudgeSigHi] on real and \WikiFictJudgeSig{} [\WikiFictJudgeSigLo, \WikiFictJudgeSigHi] on
synthetic entities, and it ranks \WikiRealDJudgeRankSigma{} by the point estimate of selectivity, so the two
rankings differ (Figure~\ref{fig:rank}). The rank is a point estimate: the judge's interval overlaps
those of M0-last, M1, M3 and M2, and the judge and M0-last are nearly tied, so the ordering at the top
is not established by these data.

The fourth pattern, the H1 criterion, fails. Trained readouts keep large edit effects (lower bounds
above 0.25 on real entities, with one exception on synthetic entities, where the lower bound of M1 is \WikiFictMOneEeLo{}), but on
2Wiki they also show large chain effects, between \WikiRealDTrainedCEMin{} and \WikiRealDTrainedCEMax{} on
real entities, and \WikiRealDTrainedSigmaPosCount{} of the \WikiRealDNTrained{} systems covered by H1 (every trained system except M0-last) have positive point estimates of selectivity. Edit
sensitivity replicates; edit dominance does not. The same MuSiQue-trained readouts that were edit-dominated on
MuSiQue show much stronger chain sensitivity on 2Wiki. We do not know why. One of several possible
explanations is surface leakage through the title: the 2Wiki later-hop passage is titled with the bridge
entity (Section~\ref{sec:construction}), so in D the earlier passage names $X$ while the later title
names $Y$, a mismatch visible without following the chain. Entity-linking noise and the shift to
eight-passage sets and 2Wiki question types are other candidates. A follow-up title-masking control finds
the title to be a small cue for the readouts and not detectably one for the judge; the other candidates
are untested (Section~\ref{sec:open}).

\begin{figure}[!htbp]
\centering
\includegraphics[width=\textwidth]{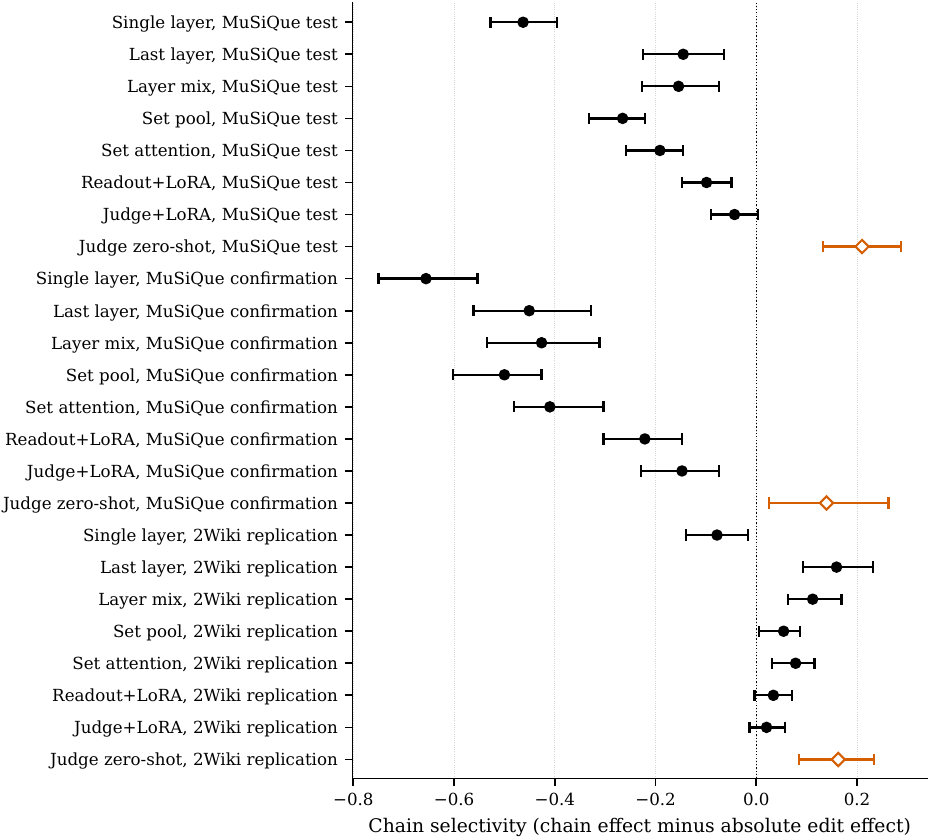}
\caption{Chain selectivity with 95\% pair-bootstrap intervals in all three settings, real bridge entities.
Open diamonds mark the reference zero-shot judge. A system is chain-selective when its interval lies
right of the dotted zero line. The synthetic-entity version is Figure~\ref{fig:fict}.}
\label{fig:forest}
\end{figure}

\subsection{Order and construction diagnostics}
Table~\ref{tab:diag} collects both diagnostics on the MuSiQue test split.
\begin{table}[t]
\centering\small
\begin{tabular}{lccc}
\toprule
System & Order jitter ratio & Hop-removal transfer & ABSENT transfer \\
\midrule
M0 single layer & \GeMZero & \TrRemovalMZero & \TrAbsentMZero \\
M0-last & \GeMZerolast & \TrRemovalMZerolast & \TrAbsentMZerolast \\
M1 layer mix & \GeMOne & \TrRemovalMOne & \TrAbsentMOne \\
M2 set pool & \GeMTwo & \TrRemovalMTwo & \TrAbsentMTwo \\
M3 set attention & \GeMThree & \TrRemovalMThree & \TrAbsentMThree \\
M4-R readout+LoRA & \GeMFoure & \TrRemovalMFoure & \TrAbsentMFoure \\
M4-J judge+LoRA & \GeMFourj & \TrRemovalMFourj & \TrAbsentMFourj \\
Judge zero-shot & \GeJudge & \TrRemovalJudge & \TrAbsentJudge \\
\bottomrule
\end{tabular}
\caption{Diagnostics on the MuSiQue test split. Order jitter ratio: mean per-record score standard deviation over three re-extracted passage orders divided by the mean FULL-to-PARTIAL gap; the pre-specified pass criterion is below 0.25. Transfer: nominal pair AUC of systems trained on substitution, scored on sets built by another method. }
\label{tab:diag}
\end{table}

Under re-extraction with
shuffled passage order the single-vector probes fail the pre-specified jitter criterion, and the set
models, the adapted systems and the zero-shot judge pass. The backbone is causal, so set models are not
order-invariant by construction; this control has to be measured on re-extracted features. Construction
transfer does not track selectivity: M1 has negative selectivity on MuSiQue but the highest hop-removal
transfer among trained systems. Hop removal also has a high surface ceiling (Table~\ref{tab:surface}), so
transfer there may reflect query-relevance cues, and we treat this axis as a diagnostic only.

\FloatBarrier
\section{Post-hoc robustness analysis}\label{sec:posthoc}
Nothing in this section was pre-specified. It reuses the frozen models, scores and pairs, uses
$B=\PhBootB$ resamples with a fixed seed, and changes no number or verdict reported above. We report it
because it determines how far the pre-specified results generalise.

\paragraph{Reproduction.} Recomputing the MuSiQue confirmation metrics from the frozen features failed
our strict reproduction gate. The nominal score of M0 came out as \PhMuMZeroNomNow{} against the ledger
value \PhMuMZeroNomLedger{}. M0 produces \PhMuMZeroTies{} exact A/D score ties, and \PhMuMZeroBitequal{}
of its pairs have bit-identical scores. A single-threaded BLAS rerun returns the ledger value
(\PhMuMZeroSingleNom{}) but changes the order of \PhMuMZeroFlips{} pairs; we do not assert a cause. Under
a post-hoc rule that excludes the nominal score, the reproduction is partial: the chain effect, the edit
effect, selectivity and the verdicts derived from them reproduce. The MuSiQue nominal values in this paper
are therefore ledger values from the original run and are sensitive to ties; the conditional contrasts,
which carry the claims, are not affected. The 2Wiki recomputation passed the gate.

\paragraph{Bootstrap design.} The pre-specified intervals resample pairs independently within each slice.
Pairs built from the same source question or sharing documents are not independent. We group records into
connected components of shared documents and resample whole groups, jointly across the real and synthetic
slices, taking the median over the system's scoring seeds within each resample; a coarser grouping by
passage title gives a second variant. The re-implemented unclustered bootstrap used as the reference for these
comparisons is not byte-identical to the pre-specified one: for the judge on real entities it gives
[\PhMuPairRealSigLo, \PhMuPairRealSigHi] on MuSiQue and [\PhWikiPairRealSigLo, \PhWikiPairRealSigHi] on
2Wiki, against the pre-specified [\MuRealJudgeSigLo, \MuRealJudgeSigHi] and [\WikiRealJudgeSigLo,
\WikiRealJudgeSigHi]. Verdicts use the pre-specified intervals.

\paragraph{The effective-sample problem.} A clustered interval is informative only if the groups are many
and balanced. With group sizes $n_g$, we use the effective number of groups $K_{\mathrm{eff}} =
(\sum_g n_g)^2/\sum_g n_g^2$ and call a clustered interval informative when the largest group holds at
most half the records, $K_{\mathrm{eff}}\ge\PhKeffMin$, and each slice has at least \PhPerVariantMin{} groups. Only the
largest-group rule existed before the first clustered run; the other two were added afterwards. The
MuSiQue confirmation records form \PhMuNclusters{} document groups (\PhMuNquads{} records, largest group
\PhMuLargest\%), with $K_{\mathrm{eff}}=\PhMuKeff$; grouping by title gives $K_{\mathrm{eff}}=\PhMuTitleKeff$.
The 2Wiki records form \PhWikiNclusters{} groups (\PhWikiNquads{} records), the largest holding
\PhWikiLargest\% of them. Neither dataset yields an informative clustered interval.

\paragraph{How we read it.} We apply three rules. First, the pre-specified pair-bootstrap verdicts stand as
reported; a post-hoc analysis does not turn them into failures. Second, the $K_{\mathrm{eff}}$ rule is post
hoc, so we do not use it to rescue H2 either: we do not claim that H2 survives clustering. Third, an
uninformative clustered interval is not a refutation. On MuSiQue the clustered interval for the judge's
selectivity on real entities is [\PhMuClusterRealSigLo, \PhMuClusterRealSigHi] and includes zero, while on
synthetic entities it is [\PhMuClusterFictSigLo, \PhMuClusterFictSigHi]; on 2Wiki the corresponding
intervals are [\PhWikiClusterRealSigLo, \PhWikiClusterRealSigHi] and [\PhWikiClusterFictSigLo,
\PhWikiClusterFictSigHi]. None of these is decisive, in either direction. H2 therefore holds as a
pre-specified criterion met on these records; it cannot be read as general confirmation that the judge is
chain-selective.

The H1 and H3 results are a separate matter from informativeness. On MuSiQue, H1 holds under the pair
bootstrap and under both groupings, but it remains a within-distribution check on documents shared with training. The
judge's real-minus-synthetic edit-effect difference (H3) is \PhMuClusterHThree{} [\PhMuClusterHThreeLo,
\PhMuClusterHThreeHi] under document clustering on MuSiQue and \PhWikiClusterHThree{}
[\PhWikiClusterHThreeLo, \PhWikiClusterHThreeHi] on 2Wiki; because the two slices differ in membership,
H3 stays exploratory. Finally, the judge ranks first by point-estimate selectivity on both datasets, but
the rank means different things. On 2Wiki it rests on a difference from M0-last of \PhWikiPairJm{}
[\PhWikiPairJmLo, \PhWikiPairJmHi] under the pair bootstrap, a near-tie; on MuSiQue the same difference is
\PhMuPairJm{} [\PhMuPairJmLo, \PhMuPairJmHi], against a readout that is edit-dominated there. Which system
ranks first, and by how much, is a dataset-specific point estimate rather than a general ordering.

\section{Discussion}

\paragraph{Two contrasts.} High edit sensitivity does not imply the absence of a positive conditional chain contrast. On
MuSiQue the trained readouts are edit-dominated; on 2Wiki the same MuSiQue-trained readouts are both edit- and
chain-sensitive. An ordinary score mixes the two, and so does any benchmark that reports nominal
accuracy alone. One hypothesis for the dataset dependence is that the 2Wiki later-hop passage is titled
with the bridge entity, so the substitution also changes the passage title and makes the break visible at
the surface; a pre-specified follow-up title-masking control tested it: the observed shift is a few hundredths of
AUC for the trained readouts and not detectable for the judge at the record level; because its verdict is
inconclusive under document dependence, this does not rule out title-related explanations beyond these
records (Section~\ref{sec:open}).

\paragraph{Construction and dependence as findings.} Two results of this study are about the protocol
rather than about any system. First, on the document-dependent MuSiQue confirmation records, the edit-matched
control changes the point estimate of M0's chain contrast from negative to positive. This shows
construction sensitivity on these records; because the verdict is INCONCLUSIVE\_DEPENDENCE, its generality
is unresolved, and a sign read from one construction should not be taken as a property of the system alone. Second, the reliability of every comparison here is
bounded by document dependence, not by the number of records or systems: 2Wiki has fewer than two
effective document groups in every slice, and only one MuSiQue slice reaches the informativeness
threshold. Both belong in the report of any edited sufficiency benchmark.

\paragraph{Judge families and datasets.} The additional judges argue against a simple reading in which
zero-shot judges are inherently better at following evidence chains. Only \MuDExtraJudgesPass{} of
\MuDExtraJudgesN{} is chain-selective on the MuSiQue confirmation set, yet all \WikiDExtraJudgesPass{}
of \WikiDExtraJudgesN{} are on 2Wiki. Chain and edit sensitivity vary with the system
family and with the dataset, and the nominal score shows neither.

\paragraph{Lessons for benchmark builders.} Ship an edit-only cell with any construction that edits text;
without it, a score near the ceiling cannot be attributed to chain following, and it cannot be dismissed
as an artifact either. Match the amount of editing between the control and treatment cells; where that is
not possible, as here, state which direction the mismatch biases each contrast. Validate every cell in code,
including the control cells: our own consistent swap broke query-lexical neutrality in
\TestRealQEight{} of \TestRealNcand{} real-entity test candidates. Use synthetic entities before reading a residual edit
effect as a weakness. Group splits by document, and check that the group key means what you think it does. Record the batch
layout of every scoring run, and keep it fixed when cells scored in different runs are compared.

\section{Limitations}
\begin{enumerate}
\item \emph{Evidence tiers.} Section~\ref{sec:explore} is exploratory. The confirmation pairs come from
the same source dataset and share documents with training, so confirmation rests on the untrained judge
and on contrasts. The 2Wiki replication reproduced three of four patterns; edit dominance of trained
readouts is not part of our claim. Section~\ref{sec:hotpot} adds one confirmatory result, which is a
non-replication, one exploratory result and one indicative result; the eligibility rule of its
confirmation set was revised after a manifest check and before any score.
\item \emph{Thin margin and dependence.} H2 passed on MuSiQue real entities with a lower bound of
\MuRealJudgeSigLo. Records share documents, and too few independent document groups exist for an
informative clustered interval on either dataset (Section~\ref{sec:posthoc}), so H2 is not general
confirmation of chain selectivity. On the independent HotpotQA set, where clustered intervals are
informative, the H2-type pattern did not replicate (Section~\ref{sec:hotpotconf}).
\item \emph{Rank stability and multiple comparisons.} Rankings are point-estimate ranks of eight systems
whose selectivity intervals overlap, and only three hypotheses were pre-specified; every other contrast
across systems, slices and judge families is reported without correction for their number. Intervals are
pair bootstraps; pairs built from the same source question or sharing documents are not independent.
A post-hoc cluster bootstrap over document groups widens the intervals and is not informative on these
data (Section~\ref{sec:posthoc}).
\item \emph{Partial reproduction.} Recomputed from frozen features, the MuSiQue conditional contrasts and
their verdicts reproduce, but the MuSiQue nominal score of M0 is numerically unstable under recomputation
(\PhMuMZeroNomNow{} against the ledger value \PhMuMZeroNomLedger{}) in a case with many exact score ties;
the cause is unresolved, and the nominal rank of M0 inherits this instability (Section~\ref{sec:posthoc}).
\item \emph{Partial synthetic control.} Only the bridge entity is renamed; other entities keep their
real names.
\item \emph{Scope.} One backbone (Qwen3.8-27B), one language (English), three source datasets (HotpotQA only in Section~\ref{sec:hotpot}). All eight
primary systems share that checkpoint, so the rank contrast is a within-backbone result; the additional
judges come from other families but are untrained, so no cross-backbone comparison of trained readouts
exists. 2Wiki
carries entity-linking noise (a later-hop page may describe a namesake); labels still follow from
provenance, but such passages are less natural.
\item \emph{Missing cell.} C is not constructed, so no cell isolates a broken chain without an edit; CE
and EE are conditional contrasts, no interaction is estimable, and B and D differ in the number of edited
passages (two versus one). The crossed control of Section~\ref{sec:crossed} cancels additive
per-passage effects but does not construct C either.
\item \emph{Validator scope.} Q1 to Q9 certify string- and graph-level properties. They do not certify the
naturalness or semantic plausibility of an edited passage, and no human or blinded model audit of the
edited text was run, so a system may be reacting to unnatural text rather than to an edit as such; the
protocol counts both as edit trace.
\item \emph{Selection of independent groups.} A confirmation set built only from records that share no
document, question or group with each other or with any earlier split removes frequently reused documents.
Independence then comes at the cost of representativeness: the retained items may differ in distribution
from the source datasets, so results on such a set describe that set, not the full benchmark population. A first
attempt under a separately fixed eligibility protocol produced no slice that met its minimum
independent-sample threshold and contributes no result. The HotpotQA set of Section~\ref{sec:hotpotconf}
meets the threshold only after a post-manifest amendment of its eligibility rule and comes from the
training split of one source dataset, so its non-replication describes that set.
\item \emph{Batch layout.} Judge scores depend on how records are packed into batches
(Section~\ref{sec:batch}). Our results are reproducible under the recorded layouts, not invariant to
batching, and the cause of the largest single-record shift is unexplained.
\item \emph{Closed-book filter.} The rule is lenient string containment; residual parametric leakage is
not ruled out, and the filter uses the reference backbone only.
\end{enumerate}

\section{Pre-specified follow-up controls}\label{sec:open}
\paragraph{Status.} The two controls below were designed after the original MuSiQue and 2Wiki results
had been seen. Their samples, statistics and decision rules were fixed in an internal record before any
of their scores existed, and auxiliary analyses were added to that record, again before any score was
read, without changing a decision rule. They are pre-specified follow-up controls, not a preregistration
of the study. They test two alternative explanations of the judge's selectivity; neither is meant to
make H2 pass. All four verdicts (two controls, two datasets) are INCONCLUSIVE\_DEPENDENCE under the
dependence rule of Section~\ref{sec:posthoc}, which is what the group structure predicted before any
score: on MuSiQue the synthetic slice and the joint comparison have fewer than \PhKeffMin{} effective
clusters, and on 2Wiki one document group carries most records in every slice
(Table~\ref{tab:ffkeff}). We therefore report effect sizes rather than verdicts: the direction, the
point change, a record-level interval and a cluster interval, with the effective cluster count. The
primary comparison is with the original scores on the same records: one paired bootstrap
(\PhBootB{} resamples) draws each record's original and controlled cells together, so the change has
its own interval. Results on the full original hard set are reported in the analysis output as
auxiliary only. Intervals on 2Wiki rest on fewer than two effective groups and should be read that way.

\begin{table}[t]
\centering
\small
\begin{tabular}{llrrrr}
\toprule
Control & Dataset & Slice & Records & Clusters & $K_{\mathrm{eff}}$ (largest share, \%) \\
\midrule
Edit-matched & MuSiQue & real & \FeOneMuRealNq & \FeOneMuRealNcl & \FeOneMuRealKeff{} (\FeOneMuRealLargest) \\
 & & synthetic & \FeOneMuFictNq & \FeOneMuFictNcl & \FeOneMuFictKeff{} (\FeOneMuFictLargest) \\
 & & joint & \FeOneMuJointNq & \FeOneMuJointNcl & \FeOneMuJointKeff{} (\FeOneMuJointLargest) \\
 & 2Wiki & real & \FeOneWikiRealNq & \FeOneWikiRealNcl & \FeOneWikiRealKeff{} (\FeOneWikiRealLargest) \\
 & & synthetic & \FeOneWikiFictNq & \FeOneWikiFictNcl & \FeOneWikiFictKeff{} (\FeOneWikiFictLargest) \\
 & & joint & \FeOneWikiJointNq & \FeOneWikiJointNcl & \FeOneWikiJointKeff{} (\FeOneWikiJointLargest) \\
\midrule
Title masking & MuSiQue & real & \FeTwoMuRealNq & \FeTwoMuRealNcl & \FeTwoMuRealKeff{} (\FeTwoMuRealLargest) \\
 & & synthetic & \FeTwoMuFictNq & \FeTwoMuFictNcl & \FeTwoMuFictKeff{} (\FeTwoMuFictLargest) \\
 & & joint & \FeTwoMuJointNq & \FeTwoMuJointNcl & \FeTwoMuJointKeff{} (\FeTwoMuJointLargest) \\
 & 2Wiki & real & \FeTwoWikiRealNq & \FeTwoWikiRealNcl & \FeTwoWikiRealKeff{} (\FeTwoWikiRealLargest) \\
 & & synthetic & \FeTwoWikiFictNq & \FeTwoWikiFictNcl & \FeTwoWikiFictKeff{} (\FeTwoWikiFictLargest) \\
 & & joint & \FeTwoWikiJointNq & \FeTwoWikiJointNcl & \FeTwoWikiJointKeff{} (\FeTwoWikiJointLargest) \\
\bottomrule
\end{tabular}
\caption{Dependence structure of the follow-up controls. Clusters are shared-document groups;
$K_{\mathrm{eff}}$ is the effective number of clusters and the parenthesis gives the share of records in
the largest cluster. A clustered interval counts as informative only with at least \PhKeffMin{}
effective clusters, at least \PhKeffMin{} clusters and a largest share of at most one half. Only the
MuSiQue real slice of the edit-matched control meets the rule; the title-masking MuSiQue real slice
falls just below it after masking excludes records.}
\label{tab:ffkeff}
\end{table}

\paragraph{Control 1: edit matching.} The new cell $D'$ edits the earlier passage as well
($X\to Z$). On 2Wiki real, $D'$ could not be built for every pair because no second substitute entity $Z$
was available for some of them, so this control uses \FeOneWikiRealNq{} of the \WikiRealNhard{} real pairs;
the MuSiQue slices and the 2Wiki synthetic slice keep every pair. The control scores six systems: the
judge, M0, M0-last, M1, M2 and M3. With $D'$, the chain contrast $\text{CE}_m=\mathrm{AUC}(B>D')-0.5$ compares two edited passages with two.
The asymmetry predicted in Section~\ref{sec:protocol} is visible: for a scorer that counts edits, the
original CE should rise once edits are matched. It rises for every system on MuSiQue and falls for none on 2Wiki. For the judge the change
on MuSiQue is \FeOneMuRealJDce{} on the real slice (record interval [\FeOneMuRealJDcePlo, \FeOneMuRealJDcePhi],
cluster [\FeOneMuRealJDceClo, \FeOneMuRealJDceChi]) and \FeOneMuFictJDce{} on the synthetic slice
([\FeOneMuFictJDcePlo, \FeOneMuFictJDcePhi]; cluster [\FeOneMuFictJDceClo, \FeOneMuFictJDceChi]); on 2Wiki it
is \FeOneWikiRealJDce{} (real, [\FeOneWikiRealJDcePlo, \FeOneWikiRealJDcePhi]; cluster
[\FeOneWikiRealJDceClo, \FeOneWikiRealJDceChi]) and \FeOneWikiFictJDce{} (synthetic,
[\FeOneWikiFictJDcePlo, \FeOneWikiFictJDcePhi]). The edit-dominated readout M0 moves far more on MuSiQue:
its CE goes from \FeOneMuRealMzeroCeo{} to \FeOneMuRealMzeroCem{} on the real slice and from
\FeOneMuFictMzeroCeo{} to \FeOneMuFictMzeroCem{} on the synthetic slice, a reversal of sign on both
slices. This is the largest change either control produced. It shows that the negative chain contrasts
of the trained readouts on MuSiQue are highly sensitive to the edit-matching intervention; it does not
yet show that the edit-count asymmetry alone caused them, because matching the edit count also edits a
second passage and may change its coherence, naturalness or other surface cues. The automated
artifact-detectability audit below is designed to separate these in part, and the reversal has so far been observed only on the
document groups of this confirmation set.

The point estimate of $\Sigma$ remains positive after matching in all four analyzed slices. The judge's $\Sigma_m$ stays positive:
\FeOneMuRealJSigm{} on MuSiQue real (record [\FeOneMuRealJSigmPlo, \FeOneMuRealJSigmPhi], cluster
[\FeOneMuRealJSigmClo, \FeOneMuRealJSigmChi]), the one slice whose clustered interval is informative, and
\FeOneMuFictJSigm{}, \FeOneWikiRealJSigm{} and \FeOneWikiFictJSigm{} elsewhere, with every record and cluster
lower bound above zero. On the MuSiQue real slice every trained readout stays edit-dominated, with record intervals below zero; M0 has $\Sigma_m$ of \FeOneMuRealMzeroSigm{}
(real, cluster [\FeOneMuRealMzeroSigmClo, \FeOneMuRealMzeroSigmChi]) and \FeOneMuFictMzeroSigm{}
(synthetic, cluster [\FeOneMuFictMzeroSigmClo, \FeOneMuFictMzeroSigmChi]), down from \FeOneMuRealMzeroSigo{}
and \FeOneMuFictMzeroSigo{}. On the synthetic slice two of the five readouts no longer separate from
zero, and on 2Wiki M0's near-zero $\Sigma$ turns slightly positive (\FeOneWikiRealMzeroSigo{} to
\FeOneWikiRealMzeroSigm{}, cluster [\FeOneWikiRealMzeroSigmClo, \FeOneWikiRealMzeroSigmChi]). Edit matching
therefore changes the size of both contrasts, by a lot for the readouts, and moves readouts that were
close to the boundary. Only the MuSiQue real slice has an informative clustered interval, and every
control verdict is INCONCLUSIVE\_DEPENDENCE, so these estimates do not establish that the judge's chain
selectivity or the readouts' edit dominance is unchanged. The judge ranks
first by $\Sigma$ before and after matching in every slice; the rank correlation of these six systems
between $\Sigma$ and $\Sigma_m$ is \FeOneMuRealRho{} (MuSiQue real), \FeOneMuFictRho{} (MuSiQue
synthetic), \FeOneWikiRealRho{} (2Wiki real) and \FeOneWikiFictRho{} (2Wiki synthetic). One asymmetry
remains by construction: $B$ introduces one new entity and $D'$ two.

\paragraph{Control 2: title masking.} Every passage title in every cell was replaced by a neutral placeholder,
and an unmasked copy of the same cells was scored in the same run. Records whose bridge entity no longer
appears in any passage body after masking were excluded, as specified: 2Wiki keeps \FeTwoWikiRealNkept{}
of \FeTwoWikiRealNhard{} real and \FeTwoWikiFictNkept{} of \FeTwoWikiFictNhard{} synthetic records,
MuSiQue \FeTwoMuRealNkept{} of \FeTwoMuRealNhard{} and \FeTwoMuFictNkept{} of \FeTwoMuFictNhard{}. The masking
changed what it was meant to change and little else. On 2Wiki the bridge entity named the earlier-hop
title in \FeTwoWikiRealAXintitleu\% of real $A$ cells and the substituted entity named the later-hop title
in \FeTwoWikiRealDYintitleu\% of real $D$ cells; after masking both are \FeTwoWikiRealDYintitlem\%, while the
bridge entity stays in the $D$ body in \FeTwoWikiRealDXinbodym\% of records. Length fell by the same small
amount in every cell, from \FeTwoWikiRealDToku{} to \FeTwoWikiRealDTokm{} tokens for the real $D$ cell on
2Wiki and from \FeTwoMuRealDToku{} to \FeTwoMuRealDTokm{} on MuSiQue, and the passage count, order and body text
are unchanged by construction. MuSiQue is a weaker negative control than intended: its earlier-hop title
names the bridge entity in \FeTwoMuRealAXintitleu\% of real $A$ cells.

Masking leaves the judge's chain contrast on 2Wiki almost unchanged: $\Delta_{\text{mask}}=\text{CE}_M-\text{CE}_U$
is \FeTwoWikiRealJDmask{} on the real slice (record [\FeTwoWikiRealJDmaskPlo, \FeTwoWikiRealJDmaskPhi], cluster
[\FeTwoWikiRealJDmaskClo, \FeTwoWikiRealJDmaskChi]) and \FeTwoWikiFictJDmask{} on the synthetic slice
([\FeTwoWikiFictJDmaskPlo, \FeTwoWikiFictJDmaskPhi]; cluster [\FeTwoWikiFictJDmaskClo, \FeTwoWikiFictJDmaskChi]).
Both record intervals lie inside the pre-specified equivalence margin of $\pm$\FeTwoMargin{} AUC; the
synthetic cluster interval does not, and no 2Wiki cluster interval is informative. Against the original
scores on the same records the masked CE differs by \FeTwoWikiRealJDce{} (real, [\FeTwoWikiRealJDcePlo,
\FeTwoWikiRealJDcePhi]) and \FeTwoWikiFictJDce{} (synthetic, [\FeTwoWikiFictJDcePlo, \FeTwoWikiFictJDcePhi]),
and the judge's $\Sigma_M$ stays at \FeTwoWikiRealJSigm{} and \FeTwoWikiFictJSigm{} with record and cluster
lower bounds above zero. On MuSiQue the judge moves by a similar amount, \FeTwoMuRealJDmask{} (real) and
\FeTwoMuFictJDmask{} (synthetic), so the pre-specified reading that 2Wiki should move more than the negative
control does not hold for the judge. That small shifts of this size are within run-to-run variation is
suggested by the unmasked rebuild itself: on MuSiQue real its CE is \FeTwoMuRealJCeu{} against
\FeTwoMuRealJCeo{} for the original cells.

The trained readouts behave differently. On 2Wiki real, masking lowers CE for every readout, by
\FeTwoWikiRealMzeroDmask{} for M0 (record [\FeTwoWikiRealMzeroDmaskPlo, \FeTwoWikiRealMzeroDmaskPhi]) and
\FeTwoWikiRealMthreeDmask{} for M3 ([\FeTwoWikiRealMthreeDmaskPlo, \FeTwoWikiRealMthreeDmaskPhi]), whereas on
MuSiQue the same readouts show no consistent direction (M0 \FeTwoMuRealMzeroDmask{}, M3
\FeTwoMuRealMthreeDmask{}). The title is a measurable but small channel for the readouts on 2Wiki: it
explains a few hundredths of AUC, not their large 2Wiki chain contrasts. System ranks by $\Sigma$ are
nearly unchanged by masking (rank correlation with the original \FeTwoWikiRealRhom{} on 2Wiki real and
\FeTwoWikiFictRhom{} on 2Wiki synthetic); the judge is first after masking in every slice, and second only
in the unmasked 2Wiki real rebuild.

\paragraph{What the controls change.} Neither control returns a decisive verdict, and neither is used to
change H1, H2 or H3. Read as effect sizes they narrow the explanations. The edit-count asymmetry is a
construction variable that moves every system's chain contrast in the predicted direction and moves the
trained readouts most, but it does not create the judge's positive $\Sigma$ and does not remove the
readouts' edit dominance on the MuSiQue real slice; readouts near the boundary move across it. Masking the bridge entity in the 2Wiki title moves the trained readouts by a few
hundredths and has no detectable record-level effect on the judge; because the verdict is inconclusive,
this does not rule out title-related explanations of the large 2Wiki chain contrasts. The dataset dependence of trained-readout edit dominance---the H1 pattern that failed on 2Wiki---remains unexplained. The shift in set size and question type between the two
datasets is the first untested candidate we would examine, with entity-linking noise as another; neither
has been tested.

\paragraph{Open controls.} Two of the controls planned at this point have since been run
(Section~\ref{sec:hotpot}): a confirmation set of independent document groups and, within the crossed
control, an artifact-only edit detector cross-fitted over document groups together with model-based
naturalness ratings by judges outside the backbone family. These do not establish human-validated semantic
plausibility, and detectability does not show that a system used the cue. Still unrun is a trained readout
on a second backbone, which would show whether the within-backbone rank contrast is specific to this
checkpoint; scored on the same dependent records it would not narrow the clustered intervals, so it
belongs on independent sets.

\section{Independent confirmation and a crossed edit-trace control}\label{sec:hotpot}

\paragraph{Status and evidence tiers.} The analyses in this section were run after every result above
had been read, to address the limitation that ends Section~\ref{sec:posthoc}: no earlier slice had enough
independent document groups for an informative clustered interval. Each was fixed in an internal record
before its scores existed; none is a public preregistration, and none changes H1, H2, H3 or any verdict
above. The four results carry different weight, and we report them at the tier at which they were
designed (Table~\ref{tab:tiers}).

\begin{table}[t]
\centering\small
\begin{tabular}{>{\raggedright\arraybackslash}p{.22\textwidth}>{\raggedright\arraybackslash}p{.15\textwidth}>{\raggedright\arraybackslash}p{.37\textwidth}l}
\toprule
Analysis & Records & Finding & Evidence grade \\
\midrule
HotpotQA independent confirmation & \HpN{} pairs & H1-type and H2-type patterns not replicated & confirmatory \\
Crossed control, arm E & \CxNE{} pairs & all \CxNCs{} systems bridge-consistency sensitive (CS4); all edit-count sensitive & exploratory \\
Crossed control, holdouts C1/C2 & \CxNCone{} / \CxNCtwo{} pairs & CS4 direction agrees for every system & indicative \\
Batch-layout diagnostic & \BxN{} records & layout dependence confirmed; gate passes after recovery & methodological \\
\bottomrule
\end{tabular}
\caption{Evidence tiers of the analyses in Section~\ref{sec:hotpot}. Confirmatory: decision rules, sample and
systems fixed before any score of the set was computed. Exploratory: items whose earlier scores had been
read. Indicative: unseen holdouts too small for a verdict under the pre-specified size rule, reported as point
estimates only. Methodological: a reproducibility check of the scoring pipeline, not a result about systems.
The grade describes how the analysis was designed, not how large an effect is; effect sizes and intervals are
in Tables~\ref{tab:hotpot} and~\ref{tab:crossed}.}
\label{tab:tiers}
\end{table}

\subsection{HotpotQA independent confirmation}\label{sec:hotpotconf}
\paragraph{Design.} We built the same A/B/D/D$'$ cells from the HotpotQA training split, which none of
the systems saw in training or selection. Every record was checked against every training, selection,
control and synthetic manifest used earlier (question identifiers, normalised gold-passage titles and
passage-body hashes). The first check failed on gold-title overlap, and the set was stopped. A revised
eligibility rule---drop every pair any of whose gold passages shares a normalised title with a passage of
those manifests---was then fixed after seeing that check and before any system score, and is reported as a
post-manifest amendment, not as the original design. It kept \HpAfter{} of \HpBefore{} real-entity pairs,
all of which passed the full check again; the closed-book filter of Section~\ref{sec:construction} left
\HpN{} pairs. Under the document-group clustering of Section~\ref{sec:posthoc} every pair is its own
group ($K_{\mathrm{eff}}=\HpKeff$), so, unlike every earlier slice, clustered and pair intervals coincide
and are informative. Systems, checkpoints, prompts, metrics and the thresholds of H1 and H2 are those of
Section~\ref{sec:confirm}, applied without retraining; the trained readouts and LoRA systems were trained
on MuSiQue, so this is also a cross-dataset test for them. Synthetic-entity pairs were scored as a
diagnostic only.

\paragraph{Result.} Neither pattern replicated (Table~\ref{tab:hotpot}). The H1-type pattern (edit effect
lower bound above 0.25 and selectivity upper bound below zero for every trained system) held for
\HpHOnePass{} of the \HpHOneN{} trained systems; M1 is not edit-dominated (selectivity upper bound
\HpMoneSighi{}). The H2-type pattern failed for the reference judge, whose selectivity lower bound is
\HpJudgeSiglo{}. Of the \HpXjN{} additional judges, \HpXjPass{} have a selectivity interval above zero and
phi-4 does not (lower bound \HpPhiSiglo{}). This is the only set in this study on which the dependence rule
allows a decisive clustered verdict, and on it the pre-specified pattern does not hold. We therefore do not
read the MuSiQue and 2Wiki passes of H2 as transferring to an independent sample. With the edit-matched
cell $D'$ in place of $D$, the trained readouts' $\Sigma_m$ lies between \HpSigmMin{} and \HpSigmMax{}, so
on these records much of their negative selectivity tracks the unequal number of edited passages, as on
MuSiQue (Section~\ref{sec:open}).

\begin{table}[t]
\centering\footnotesize
\setlength{\tabcolsep}{3pt}
\begin{tabular}{lccccc}
\toprule
System & Nominal & CE & EE & $\Sigma$ [95\% CI] & $\Sigma_m$ [95\% CI] \\
\midrule
M0 single layer & \HpMZeroNom & \HpMZeroCe & \HpMZeroEe & \HpMZeroSig{} [\HpMZeroSigLo, \HpMZeroSigHi] & \HpMZeroSigm{} [\HpMZeroSigmLo, \HpMZeroSigmHi] \\
M0-last & \HpMZerolastNom & \HpMZerolastCe & \HpMZerolastEe & \HpMZerolastSig{} [\HpMZerolastSigLo, \HpMZerolastSigHi] & \HpMZerolastSigm{} [\HpMZerolastSigmLo, \HpMZerolastSigmHi] \\
M1 layer mix & \HpMOneNom & \HpMOneCe & \HpMOneEe & \HpMOneSig{} [\HpMOneSigLo, \HpMOneSigHi] & \HpMOneSigm{} [\HpMOneSigmLo, \HpMOneSigmHi] \\
M2 set pool & \HpMTwoNom & \HpMTwoCe & \HpMTwoEe & \HpMTwoSig{} [\HpMTwoSigLo, \HpMTwoSigHi] & \HpMTwoSigm{} [\HpMTwoSigmLo, \HpMTwoSigmHi] \\
M3 set attention & \HpMThreeNom & \HpMThreeCe & \HpMThreeEe & \HpMThreeSig{} [\HpMThreeSigLo, \HpMThreeSigHi] & \HpMThreeSigm{} [\HpMThreeSigmLo, \HpMThreeSigmHi] \\
M4-R readout+LoRA & \HpMFoureNom & \HpMFoureCe & \HpMFoureEe & \HpMFoureSig{} [\HpMFoureSigLo, \HpMFoureSigHi] & \HpMFoureSigm{} [\HpMFoureSigmLo, \HpMFoureSigmHi] \\
M4-J judge+LoRA & \HpMFourjNom & \HpMFourjCe & \HpMFourjEe & \HpMFourjSig{} [\HpMFourjSigLo, \HpMFourjSigHi] & \HpMFourjSigm{} [\HpMFourjSigmLo, \HpMFourjSigmHi] \\
\midrule
Qwen3.8-27B judge (reference) & \HpJudgeNom & \HpJudgeCe & \HpJudgeEe & \HpJudgeSig{} [\HpJudgeSigLo, \HpJudgeSigHi] & \HpJudgeSigm{} [\HpJudgeSigmLo, \HpJudgeSigmHi] \\
Llama-3.1-8B-Instruct & \HpLlamaNom & \HpLlamaCe & \HpLlamaEe & \HpLlamaSig{} [\HpLlamaSigLo, \HpLlamaSigHi] & \HpLlamaSigm{} [\HpLlamaSigmLo, \HpLlamaSigmHi] \\
phi-4 & \HpPhiNom & \HpPhiCe & \HpPhiEe & \HpPhiSig{} [\HpPhiSigLo, \HpPhiSigHi] & \HpPhiSigm{} [\HpPhiSigmLo, \HpPhiSigmHi] \\
Qwen3-32B & \HpQwenNom & \HpQwenCe & \HpQwenEe & \HpQwenSig{} [\HpQwenSigLo, \HpQwenSigHi] & \HpQwenSigm{} [\HpQwenSigmLo, \HpQwenSigmHi] \\
Gemma-4-31B-it & \HpGemmaNom & \HpGemmaCe & \HpGemmaEe & \HpGemmaSig{} [\HpGemmaSigLo, \HpGemmaSigHi] & \HpGemmaSigm{} [\HpGemmaSigmLo, \HpGemmaSigmHi] \\
\bottomrule
\end{tabular}
\caption{HotpotQA independent confirmation (train split, real entities, closed-book-hard, $n=\HpN$ pairs, each its own document group). Intervals are document-group bootstraps. $\Sigma_m=\text{CE}_m-|\text{EE}|$ replaces $D$ with the edit-matched cell $D'$. M0-last is a reference column outside the H1-type set.}
\label{tab:hotpot}
\end{table}

\subsection{Crossed edit-trace control}\label{sec:crossed}
\paragraph{Design.} The ordinary contrasts compare cells that differ in both chain status and edit trace.
The crossed control removes additive per-passage effects. Write $e$ for the earlier supporting passage,
which mentions the bridge entity $X$, and $l$ for the later passage, titled $X$. Cell $(u,v)$ replaces $X$
by $u$ in $e$ and by $v$ in $l$; the chain is intact only when $u=v$. With a second substitute $Z$ chosen by
the same rule as $Y$, the four cells $B=(Y,Y)$, $B_Z=(Z,Z)$, $D_{ZY}=(Z,Y)$ and $D_{YZ}=(Y,Z)$ use the same
four edited passage versions, each once in an intact and once in a broken cell. The main estimand is
\[
\text{CS4}=\tfrac{1}{4}\bigl[W(B>D_{ZY})+W(B>D_{YZ})+W(B_Z>D_{ZY})+W(B_Z>D_{YZ})\bigr]-0.5,
\]
where $W$ is the paired win rate with ties counted as one half. Each of the four pairs differs in one
passage, so any scorer that adds per-passage scores, whatever those scores reflect (entity identity,
edit trace or counterfactuality), has CS4 $=0$. Two secondary contrasts isolate the edit trace: the
edit-count effect $\text{EC}=W(D>D_{ZY})-0.5$ compares two broken cells that share $l$ and differ only in
whether $e$ is also edited, and the position effect $\text{PT}=W(D_E>D)-0.5$, with $D_E=(Y,X)$, compares
editing $e$ with editing and retitling $l$. Arm E uses the crossed-eligible HotpotQA confirmation pairs
(\CxNE{} after the same closed-book filter); because their original cells had been scored, it is
exploratory by design. Two unseen holdouts, HotpotQA validation (C1) and 2WikiMultiHopQA train (C2), were
checked against all earlier sets; after that check they hold \CxNCone{} and \CxNCtwo{} pairs, below the
pre-specified size for any verdict, and serve only as a direction check. Verdicts use document-group
bootstrap intervals, Bonferroni-adjusted for the \CxFamM{} systems in the family; each statistic (CS4, EC, PT)
is corrected separately, and the artifact-only baselines below are outside the family. A system is called
\emph{chain-sensitive} (code label \texttt{CHAIN\_SENSITIVE}) when the adjusted lower bound of its CS4 is above
zero. The label is operational: it means sensitivity to bridge-string consistency, not semantic multi-hop
reasoning. Four artifact-only
baselines that never see the chain are reported beside them: three additive scorers (surface form, a
cross-fitted character n-gram edit detector, and per-passage language-model log-probability from
Llama-3.1-8B-Instruct) and a string-match reference (AB3) that scores whether the bridge string in $e$
equals the title of $l$.

\paragraph{Result.} All \CxNCs{} systems are chain-sensitive in this operational sense under the adjusted rule: CS4 ranges from
\CxCsMin{} to \CxCsMax{}, every adjusted lower bound is at least \CxCsLoMin{}, and the five judges no
longer disagree in verdict (Table~\ref{tab:crossed}). The additive baselines have CS4 $=0$ exactly, as the
estimand requires, which the analysis also checks as a validity gate. In both holdouts the sign of CS4
agrees with arm E for \CxAgreeCone{} and \CxAgreeCtwo{} of the \CxFamM{} systems. The edit trace does not
disappear. Every system also shows an edit-count effect (EC adjusted lower bound above zero for
\CxNEc{} of \CxFamM{}), and \CxNPt{} of \CxFamM{} show a position effect, all except M4-R. The artifact-only
language-model baseline reaches a nominal score of \CxAbtwoNom{} without reading the chain, so on these
records the ordinary score is available from edit traces alone. This does not contradict
Table~\ref{tab:surface}, where a surface-only classifier stays at \SurfSubstTest{} on substitution negatives:
that classifier sees query-overlap and retrieval features, which substitution leaves unchanged, whereas AB2
scores each passage with a language model and so sees the edit itself. The sets and detectors differ, so
the two numbers are not directly comparable. Edits are detectable according to model-based naturalness ratings: three non-Qwen judges, asked blind whether a single passage reads naturally, rate an edited
later passage lower than its original (Gemma-4-31B-it by \CxAudGemmaNaturalL{}, Llama-3.1-8B-Instruct by
\CxAudLlamaNaturalL{}, phi-4 by \CxAudPhiNaturalL{} in mean P(yes)), while the $Y$ and $Z$ versions are rated
alike (largest absolute difference \CxAudZyMax{}), which supports the exchangeability that CS4 assumes.
These ratings are supplementary evidence, not human validation.

\begin{table}[t]
\centering\footnotesize
\setlength{\tabcolsep}{2pt}
\begin{tabular}{lccccc}
\toprule
 & \multicolumn{3}{c}{Arm E (exploratory)} & C1 & C2 \\
\cmidrule(lr){2-4}\cmidrule(lr){5-6}
System & CS4 [adj.\ CI] & EC [adj.\ CI] & PT [adj.\ CI] & CS4 & CS4 \\
\midrule
M0 single layer & \CxMZeroCs{} [\CxMZeroCsLo, \CxMZeroCsHi] & \CxMZeroEc{} [\CxMZeroEcLo, \CxMZeroEcHi] & \CxMZeroPt{} [\CxMZeroPtLo, \CxMZeroPtHi] & \CxMZeroCone & \CxMZeroCtwo \\
M1 layer mix & \CxMOneCs{} [\CxMOneCsLo, \CxMOneCsHi] & \CxMOneEc{} [\CxMOneEcLo, \CxMOneEcHi] & \CxMOnePt{} [\CxMOnePtLo, \CxMOnePtHi] & \CxMOneCone & \CxMOneCtwo \\
M2 set pool & \CxMTwoCs{} [\CxMTwoCsLo, \CxMTwoCsHi] & \CxMTwoEc{} [\CxMTwoEcLo, \CxMTwoEcHi] & \CxMTwoPt{} [\CxMTwoPtLo, \CxMTwoPtHi] & \CxMTwoCone & \CxMTwoCtwo \\
M3 set attention & \CxMThreeCs{} [\CxMThreeCsLo, \CxMThreeCsHi] & \CxMThreeEc{} [\CxMThreeEcLo, \CxMThreeEcHi] & \CxMThreePt{} [\CxMThreePtLo, \CxMThreePtHi] & \CxMThreeCone & \CxMThreeCtwo \\
M4-R readout+LoRA & \CxMFoureCs{} [\CxMFoureCsLo, \CxMFoureCsHi] & \CxMFoureEc{} [\CxMFoureEcLo, \CxMFoureEcHi] & \CxMFourePt{} [\CxMFourePtLo, \CxMFourePtHi] & \CxMFoureCone & \CxMFoureCtwo \\
M4-J judge+LoRA & \CxMFourjCs{} [\CxMFourjCsLo, \CxMFourjCsHi] & \CxMFourjEc{} [\CxMFourjEcLo, \CxMFourjEcHi] & \CxMFourjPt{} [\CxMFourjPtLo, \CxMFourjPtHi] & \CxMFourjCone & \CxMFourjCtwo \\
\midrule
Qwen3.8-27B judge (reference) & \CxJudgeCs{} [\CxJudgeCsLo, \CxJudgeCsHi] & \CxJudgeEc{} [\CxJudgeEcLo, \CxJudgeEcHi] & \CxJudgePt{} [\CxJudgePtLo, \CxJudgePtHi] & \CxJudgeCone & \CxJudgeCtwo \\
Llama-3.1-8B-Instruct & \CxLlamaCs{} [\CxLlamaCsLo, \CxLlamaCsHi] & \CxLlamaEc{} [\CxLlamaEcLo, \CxLlamaEcHi] & \CxLlamaPt{} [\CxLlamaPtLo, \CxLlamaPtHi] & \CxLlamaCone & \CxLlamaCtwo \\
phi-4 & \CxPhiCs{} [\CxPhiCsLo, \CxPhiCsHi] & \CxPhiEc{} [\CxPhiEcLo, \CxPhiEcHi] & \CxPhiPt{} [\CxPhiPtLo, \CxPhiPtHi] & \CxPhiCone & \CxPhiCtwo \\
Qwen3-32B & \CxQwenCs{} [\CxQwenCsLo, \CxQwenCsHi] & \CxQwenEc{} [\CxQwenEcLo, \CxQwenEcHi] & \CxQwenPt{} [\CxQwenPtLo, \CxQwenPtHi] & \CxQwenCone & \CxQwenCtwo \\
Gemma-4-31B-it & \CxGemmaCs{} [\CxGemmaCsLo, \CxGemmaCsHi] & \CxGemmaEc{} [\CxGemmaEcLo, \CxGemmaEcHi] & \CxGemmaPt{} [\CxGemmaPtLo, \CxGemmaPtHi] & \CxGemmaCone & \CxGemmaCtwo \\
\midrule
AB0 surface form (additive) & \CxAbzeroCs & \CxAbzeroEc & --- & --- & --- \\
AB1 char n-gram (additive) & \CxAboneCs & \CxAboneEc & --- & --- & --- \\
AB2 LM log-prob (additive) & \CxAbtwoCs & \CxAbtwoEc & --- & --- & --- \\
AB3 string match (reference) & \CxAbthreeCs & \CxAbthreeEc & --- & --- & --- \\
\bottomrule
\end{tabular}
\caption{Crossed edit-trace control. CS4 averages four contrasts in which a chain-intact and a chain-broken cell differ in one passage and every passage version appears equally often in both chain states, so any additive per-passage scorer has CS4 $=0$; a scorer that only checks whether the bridge string in the earlier passage matches the later title (AB3) has CS4 $=0.5$. EC compares two broken cells that differ only in one extra edit; PT compares editing the earlier passage with editing (and retitling) the later one. Adjusted intervals are Bonferroni-corrected for the \CxFamM{} systems in the family (document-group bootstrap, \CxBootB{} resamples); each statistic (CS4, EC, PT) is corrected separately, and the family is the \CxFamM{} systems listed above the baselines (M0--M3, M4-R, M4-J and the five judges). Baselines are not in the family and carry no verdict. Arm E ($n=\CxNE$) reuses HotpotQA confirmation items and is exploratory; C1 ($n=\CxNCone$, HotpotQA validation) and C2 ($n=\CxNCtwo$, 2WikiMultiHopQA train) are unseen holdouts below the size threshold, reported as indicative point estimates only.}
\label{tab:crossed}
\end{table}

\paragraph{Reading.} CS4 measures sensitivity to whether the two passages name the same bridge entity
once additive per-passage effects cancel. AB3 reaches the maximum, 0.5, by checking one string, so a
positive CS4 is evidence of sensitivity to the consistency of the linking string, not of multi-hop
reasoning. We separate three claims. Sensitivity to bridge-string consistency is supported, and so is
sensitivity to the edit trace; semantic multi-hop reasoning is not established by this control. The
control therefore shows both kinds of sensitivity together in every system. It does not remove the edit confound: the edit-count effect remains, the edited passages remain
detectable, and the C cell (a broken chain without any edit) still does not exist, so non-additive
reactions such as a response to the same unfamiliar entity appearing twice are not separated from chain
consistency. Nor does it reverse the HotpotQA non-replication above: CS4 is a different estimand from
$\Sigma$, the arm is exploratory, and the holdouts are indicative.

\subsection{Batch-layout sensitivity of the scoring pipeline}\label{sec:batch}
\paragraph{What failed.} The pre-specified analysis checks that old cells rescored inside the crossed
bench reproduce their earlier scores (minimum per-cell Spearman correlation at least \BxThrRho{} and
$|\Delta W(A>D)|$ at most \BxThrDw{}). The first run failed this check for one judge, Gemma-4-31B-it
(minimum Spearman \BxRho{}), so the analysis stopped as invalid before any new-cell score was read; that
record is kept. Scoring sorts records by length and packs them into batches of at most \BxTok{} tokens, so
adding the new records changed the batches of the old ones. A diagnostic fixed before it ran showed that
identical layouts reproduce exactly in repeated runs (Spearman 1, $\Delta W=0$) and that a changed layout
alone changes scores. On the \BxN{} affected records, \BxEq\% of scores are bit-identical, the A$>$D
outcome flips in \BxFlips{} of \BxPairs{} pairs (\BxFlipfrac\%), all of them with score gaps below $10^{-3}$, and one
record moves by \BxMax{}. Large shifts of this kind are unexplained; we do not attribute them to
numerical rounding.

\paragraph{Recovery.} As fixed in the diagnostic record, we rescored the judge with the original layout:
the \BxOld{} records of the original bench as one run, and the \BxRest{} crossed records outside it as a
second bench, then merged by record identifier. The crossed set has \BxCross{} unique records,
\BxCap{} of which are in the original bench; the \BxOldonly{} original records outside the crossed set
(\BxOoFict{} synthetic-entity variants, \BxOoInel{} records of pairs that were not crossed-eligible and
\BxOoMisc{} others) were not used, and no record was selected by position or deduplicated. The merged
scores match the crossed set exactly in identifiers and order, the original-bench run reproduces the
committed scores byte for byte, and the rerun passed every validity gate (Gemma minimum Spearman
\BxAfterRho{}). This pass shows reproducibility under the recovery layout. It does not show that the
judges are invariant to batching: the other judges' old-cell scores also moved between layouts by amounts
within the threshold (minimum Spearman \BxOtherRhoMin{}, $|\Delta W|$ up to \BxOtherDwMax{}). Rankings and
win rates near ties can depend on a scoring detail that papers rarely report.

\section{Data, code and checkpoint availability}\label{sec:avail}
Every measured result in this paper is a macro generated from an experiment ledger field, and tables and
figures are generated from the same ledgers. Not all of those ledgers, records and checkpoints are public,
so external reproduction is partial. We state the boundary explicitly.

\paragraph{Public.} The assets in Table~\ref{tab:avail} are public. The ChainCheck dataset contains the
MuSiQue test and sealed-confirmation records and the 2WikiMultiHopQA replication records, together with a
standalone evaluator (\texttt{chaincheck\_eval.py}) that recomputes the paired contrasts from a score file.
The three released judges come from a separate release training run and are not the M4-R or M4-J
checkpoints evaluated here; results for M4 in this paper should not be expected from them. Our local
copies of the external judges are byte-identical to the pinned revisions.

\begin{table}[h]
\centering\small
\begin{tabular}{lll}
\toprule
Asset & Repository (Hugging Face) & Revision \\
\midrule
Dataset and evaluator & \texttt{datasets/ThakiCloud/ChainCheck} & \texttt{825d836d1b} \\
Released judges & \texttt{ThakiCloud/ChainCheck-Judge-*} & not used in this paper \\
Backbone, reference judge & \texttt{Qwen/Qwen3.8-27B} & \texttt{dbdc473dea} \\
Additional judge & \texttt{meta-llama/Llama-3.1-8B-Instruct} & \texttt{8c22764a7e} \\
Additional judge & \texttt{microsoft/phi-4} & \texttt{2db69c1c3e} \\
Additional judge & \texttt{Qwen/Qwen3-32B} & \texttt{a09d5ccb54} \\
Additional judge & \texttt{google/gemma-4-31B-it} & \texttt{842da3794e} \\
\bottomrule
\end{tabular}
\caption{Public assets. The released judges (Qwen3.5-4B, 9B and Qwen3.8-27B variants) are listed for
completeness; they are not the checkpoints evaluated in this paper.}
\label{tab:avail}
\end{table}

\paragraph{Not public.} The HotpotQA confirmation records (\HpN{} pairs), the crossed bench (arm E, C1 and
C2), the probe heads of M0--M3, the M4-R and M4-J checkpoints (identified by sha256 in internal ledgers),
the code repository with the validator and the builders, and the time-stamped pre-specification records
are internal; their release has not been approved.

\paragraph{What can be reproduced.} With the public dataset and evaluator, a reader can rescore any
system, including the pinned zero-shot judges, on the MuSiQue and 2Wiki records and recompute the
contrasts of Sections~\ref{sec:confirm} and~\ref{sec:wiki}. The trained-system results (M0--M4), the
HotpotQA confirmation, the crossed control and the batch diagnostic cannot be reproduced externally from
public material, and the pre-specification cannot be verified by a third party. We therefore do not claim
full reproducibility.

\section{Conclusion}
A near-perfect score on an edited sufficiency benchmark can conceal sensitivity to how the benchmark was
built. It does not show that a model follows evidence chains, and it does not show that the model ignores
them. ChainCheck exposes this confound by reporting, next to that score, a chain effect and an edit effect
measured as paired conditional contrasts on pairs certified by a deterministic validator. In a sealed
confirmation and an untuned replication, nominal scores stay near the ceiling while the contrasts separate
the systems: trained readouts have large edit effects, and whether edit sensitivity dominates depends on
the dataset. Under the pre-specified criterion with unclustered intervals the reference zero-shot judge is
chain-selective in every setting, but the records share documents, too few independent groups exist for
an informative clustered interval, and the judge's first rank on 2Wiki is dataset-specific. How strongly
these data support chain selectivity depends on dataset structure and on the treatment of dependent
records, which is itself a reason to report the edit-only cell and the dependence structure alongside any
sufficiency score. Two follow-up controls point the same way without settling it: matching the number of edited
passages changes every chain contrast, most for the trained readouts, and masking passage titles changes
them by a few hundredths at most; on these records the point estimates of the judge's $\Sigma$ remain
positive, but all four verdicts are inconclusive under document dependence, so neither control rules out
its alternative explanation, and neither explains why
edit dominance depends on the dataset.

On an independent HotpotQA set with one pair per document group, the only set here that allows a decisive
clustered verdict, neither pre-specified pattern replicated. A crossed control that cancels additive
per-passage effects shows that every system responds to whether the bridge-entity strings agree and that
every system also responds to the edit itself. String consistency is not multi-hop reasoning: an
artifact-only string match reaches the maximum of the same statistic. Both kinds of sensitivity are observed
together, the edit confound is not removed, and the non-replication stands. A scoring detail that papers
rarely report, how records are batched, changed one judge's scores enough to fail a reproducibility gate;
what we report is reproducibility under fixed layouts, not invariance to batching.

\FloatBarrier
\bibliographystyle{plain}
\bibliography{refs}
\clearpage

\appendix
\section{Additional tables and figures}
\begin{figure}[!htbp]
\centering
\includegraphics[width=\textwidth]{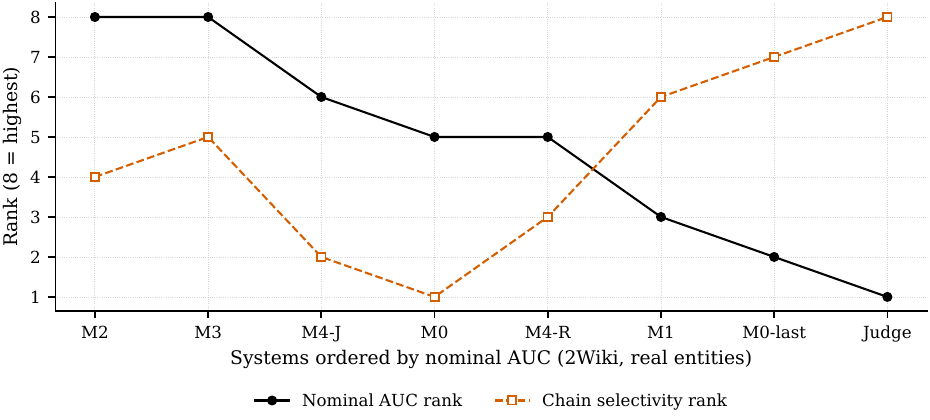}
\caption{Point-estimate rank contrast on 2WikiMultiHopQA (real bridge entities), shown for completeness. The judge and M0-last differ in selectivity by \PhWikiPairJm{} [\PhWikiPairJmLo, \PhWikiPairJmHi] under the pair bootstrap, so the rank change is not evidence of an ordering. Systems are ordered by nominal pair
AUC; solid line, rank by nominal AUC; dashed line, rank by chain selectivity. The vertical axis plots
$\WikiRealDNRankedSystems{}+1-\mathrm{rank}$, so the best system is at the top; ranks quoted in the text
use 1 for the best. M4-R and M4-J are the LoRA-adapted readout and judge. The zero-shot judge is rank
\WikiRealDJudgeRankNominal{} (last) by nominal AUC and rank \WikiRealDJudgeRankSigma{} (first) by the
point estimate of selectivity; ranks are
point-estimate ranks, not significance tests. The judge's interval (\WikiRealJudgeSig{}
[\WikiRealJudgeSigLo, \WikiRealJudgeSigHi]) overlaps those of M0-last (\WikiRealMZerolastSig{}
[\WikiRealMZerolastSigLo, \WikiRealMZerolastSigHi]), M1 (\WikiRealMOneSig{} [\WikiRealMOneSigLo,
\WikiRealMOneSigHi]), M3 (\WikiRealMThreeSig{} [\WikiRealMThreeSigLo, \WikiRealMThreeSigHi]) and M2
(\WikiRealMTwoSig{} [\WikiRealMTwoSigLo, \WikiRealMTwoSigHi]); the judge and M0-last are nearly tied, and
M2 and M3 tie by nominal AUC. Values are in Table~\ref{tab:real}.}
\label{fig:rank}
\end{figure}
\begin{table}[!ht]
\centering\small
\begin{tabular}{lcccc}
\toprule
System & Nominal & Chain effect & Edit effect & $\Sigma$ seed median [seed-avg.\ boot.] \\
\midrule
\multicolumn{5}{l}{\emph{MuSiQue test (exploratory, post hoc)}, $n=\TestRealNhard$} \\
M0 single layer & \TestRealMZeroNom & \TestRealMZeroCe & \TestRealMZeroEe & \TestRealMZeroSig{} [\TestRealMZeroSigLo, \TestRealMZeroSigHi] \\
M0-last & \TestRealMZerolastNom & \TestRealMZerolastCe & \TestRealMZerolastEe & \TestRealMZerolastSig{} [\TestRealMZerolastSigLo, \TestRealMZerolastSigHi] \\
M1 layer mix & \TestRealMOneNom & \TestRealMOneCe & \TestRealMOneEe & \TestRealMOneSig{} [\TestRealMOneSigLo, \TestRealMOneSigHi] \\
M2 set pool & \TestRealMTwoNom & \TestRealMTwoCe & \TestRealMTwoEe & \TestRealMTwoSig{} [\TestRealMTwoSigLo, \TestRealMTwoSigHi] \\
M3 set attention & \TestRealMThreeNom & \TestRealMThreeCe & \TestRealMThreeEe & \TestRealMThreeSig{} [\TestRealMThreeSigLo, \TestRealMThreeSigHi] \\
M4-R readout+LoRA & \TestRealMFoureNom & \TestRealMFoureCe & \TestRealMFoureEe & \TestRealMFoureSig{} [\TestRealMFoureSigLo, \TestRealMFoureSigHi] \\
M4-J judge+LoRA & \TestRealMFourjNom & \TestRealMFourjCe & \TestRealMFourjEe & \TestRealMFourjSig{} [\TestRealMFourjSigLo, \TestRealMFourjSigHi] \\
Judge zero-shot & \TestRealJudgeNom & \TestRealJudgeCe & \TestRealJudgeEe & \TestRealJudgeSig{} [\TestRealJudgeSigLo, \TestRealJudgeSigHi] \\
\midrule
\multicolumn{5}{l}{\emph{MuSiQue confirmation (pre-specified)}, $n=\MuRealNhard$} \\
M0 single layer & \MuRealMZeroNom & \MuRealMZeroCe & \MuRealMZeroEe & \MuRealMZeroSig{} [\MuRealMZeroSigLo, \MuRealMZeroSigHi] \\
M0-last & \MuRealMZerolastNom & \MuRealMZerolastCe & \MuRealMZerolastEe & \MuRealMZerolastSig{} [\MuRealMZerolastSigLo, \MuRealMZerolastSigHi] \\
M1 layer mix & \MuRealMOneNom & \MuRealMOneCe & \MuRealMOneEe & \MuRealMOneSig{} [\MuRealMOneSigLo, \MuRealMOneSigHi] \\
M2 set pool & \MuRealMTwoNom & \MuRealMTwoCe & \MuRealMTwoEe & \MuRealMTwoSig{} [\MuRealMTwoSigLo, \MuRealMTwoSigHi] \\
M3 set attention & \MuRealMThreeNom & \MuRealMThreeCe & \MuRealMThreeEe & \MuRealMThreeSig{} [\MuRealMThreeSigLo, \MuRealMThreeSigHi] \\
M4-R readout+LoRA & \MuRealMFoureNom & \MuRealMFoureCe & \MuRealMFoureEe & \MuRealMFoureSig{} [\MuRealMFoureSigLo, \MuRealMFoureSigHi] \\
M4-J judge+LoRA & \MuRealMFourjNom & \MuRealMFourjCe & \MuRealMFourjEe & \MuRealMFourjSig{} [\MuRealMFourjSigLo, \MuRealMFourjSigHi] \\
Judge zero-shot & \MuRealJudgeNom & \MuRealJudgeCe & \MuRealJudgeEe & \MuRealJudgeSig{} [\MuRealJudgeSigLo, \MuRealJudgeSigHi] \\
\midrule
\multicolumn{5}{l}{\emph{2WikiMultiHopQA replication (pre-specified)}, $n=\WikiRealNhard$} \\
M0 single layer & \WikiRealMZeroNom & \WikiRealMZeroCe & \WikiRealMZeroEe & \WikiRealMZeroSig{} [\WikiRealMZeroSigLo, \WikiRealMZeroSigHi] \\
M0-last & \WikiRealMZerolastNom & \WikiRealMZerolastCe & \WikiRealMZerolastEe & \WikiRealMZerolastSig{} [\WikiRealMZerolastSigLo, \WikiRealMZerolastSigHi] \\
M1 layer mix & \WikiRealMOneNom & \WikiRealMOneCe & \WikiRealMOneEe & \WikiRealMOneSig{} [\WikiRealMOneSigLo, \WikiRealMOneSigHi] \\
M2 set pool & \WikiRealMTwoNom & \WikiRealMTwoCe & \WikiRealMTwoEe & \WikiRealMTwoSig{} [\WikiRealMTwoSigLo, \WikiRealMTwoSigHi] \\
M3 set attention & \WikiRealMThreeNom & \WikiRealMThreeCe & \WikiRealMThreeEe & \WikiRealMThreeSig{} [\WikiRealMThreeSigLo, \WikiRealMThreeSigHi] \\
M4-R readout+LoRA & \WikiRealMFoureNom & \WikiRealMFoureCe & \WikiRealMFoureEe & \WikiRealMFoureSig{} [\WikiRealMFoureSigLo, \WikiRealMFoureSigHi] \\
M4-J judge+LoRA & \WikiRealMFourjNom & \WikiRealMFourjCe & \WikiRealMFourjEe & \WikiRealMFourjSig{} [\WikiRealMFourjSigLo, \WikiRealMFourjSigHi] \\
Judge zero-shot & \WikiRealJudgeNom & \WikiRealJudgeCe & \WikiRealJudgeEe & \WikiRealJudgeSig{} [\WikiRealJudgeSigLo, \WikiRealJudgeSigHi] \\
\bottomrule
\end{tabular}
\caption{All ChainCheck metrics on closed-book-hard pairs (real bridge entities). Nominal is AUC(A$>$D); chain effect is AUC(B$>$D)$-0.5$; edit effect is AUC(A$>$B)$-0.5$; selectivity is chain effect minus the absolute edit effect, with a 95\% percentile pair-bootstrap interval (2,000 resamples of accepted quads; not clustered by source question or document). For trained systems the point estimate is a median over three seeds, while the interval resamples seed-averaged per-quad outcomes, so it is not an interval for the displayed median. M0--M4 are trained on the nominal contrast; the judge is untrained. MuSiQue nominal values are ledger values from the original run; recomputed from the frozen features, the MuSiQue confirmation nominal of M0 is \PhMuMZeroNomNow{} (many exact score ties; cause unresolved; Section~\ref{sec:posthoc}), so its nominal rank is unstable.}
\label{tab:real}
\end{table}

\begin{table}[!ht]
\centering\small
\begin{tabular}{lcccc}
\toprule
System & Nominal & Chain effect & Edit effect & $\Sigma$ seed median [seed-avg.\ boot.] \\
\midrule
\multicolumn{5}{l}{\emph{MuSiQue test (exploratory, post hoc)}, $n=\TestFictNhard$} \\
M0 single layer & \TestFictMZeroNom & \TestFictMZeroCe & \TestFictMZeroEe & \TestFictMZeroSig{} [\TestFictMZeroSigLo, \TestFictMZeroSigHi] \\
M0-last & \TestFictMZerolastNom & \TestFictMZerolastCe & \TestFictMZerolastEe & \TestFictMZerolastSig{} [\TestFictMZerolastSigLo, \TestFictMZerolastSigHi] \\
M1 layer mix & \TestFictMOneNom & \TestFictMOneCe & \TestFictMOneEe & \TestFictMOneSig{} [\TestFictMOneSigLo, \TestFictMOneSigHi] \\
M2 set pool & \TestFictMTwoNom & \TestFictMTwoCe & \TestFictMTwoEe & \TestFictMTwoSig{} [\TestFictMTwoSigLo, \TestFictMTwoSigHi] \\
M3 set attention & \TestFictMThreeNom & \TestFictMThreeCe & \TestFictMThreeEe & \TestFictMThreeSig{} [\TestFictMThreeSigLo, \TestFictMThreeSigHi] \\
M4-R readout+LoRA & \TestFictMFoureNom & \TestFictMFoureCe & \TestFictMFoureEe & \TestFictMFoureSig{} [\TestFictMFoureSigLo, \TestFictMFoureSigHi] \\
M4-J judge+LoRA & \TestFictMFourjNom & \TestFictMFourjCe & \TestFictMFourjEe & \TestFictMFourjSig{} [\TestFictMFourjSigLo, \TestFictMFourjSigHi] \\
Judge zero-shot & \TestFictJudgeNom & \TestFictJudgeCe & \TestFictJudgeEe & \TestFictJudgeSig{} [\TestFictJudgeSigLo, \TestFictJudgeSigHi] \\
\midrule
\multicolumn{5}{l}{\emph{MuSiQue confirmation (pre-specified)}, $n=\MuFictNhard$} \\
M0 single layer & \MuFictMZeroNom & \MuFictMZeroCe & \MuFictMZeroEe & \MuFictMZeroSig{} [\MuFictMZeroSigLo, \MuFictMZeroSigHi] \\
M0-last & \MuFictMZerolastNom & \MuFictMZerolastCe & \MuFictMZerolastEe & \MuFictMZerolastSig{} [\MuFictMZerolastSigLo, \MuFictMZerolastSigHi] \\
M1 layer mix & \MuFictMOneNom & \MuFictMOneCe & \MuFictMOneEe & \MuFictMOneSig{} [\MuFictMOneSigLo, \MuFictMOneSigHi] \\
M2 set pool & \MuFictMTwoNom & \MuFictMTwoCe & \MuFictMTwoEe & \MuFictMTwoSig{} [\MuFictMTwoSigLo, \MuFictMTwoSigHi] \\
M3 set attention & \MuFictMThreeNom & \MuFictMThreeCe & \MuFictMThreeEe & \MuFictMThreeSig{} [\MuFictMThreeSigLo, \MuFictMThreeSigHi] \\
M4-R readout+LoRA & \MuFictMFoureNom & \MuFictMFoureCe & \MuFictMFoureEe & \MuFictMFoureSig{} [\MuFictMFoureSigLo, \MuFictMFoureSigHi] \\
M4-J judge+LoRA & \MuFictMFourjNom & \MuFictMFourjCe & \MuFictMFourjEe & \MuFictMFourjSig{} [\MuFictMFourjSigLo, \MuFictMFourjSigHi] \\
Judge zero-shot & \MuFictJudgeNom & \MuFictJudgeCe & \MuFictJudgeEe & \MuFictJudgeSig{} [\MuFictJudgeSigLo, \MuFictJudgeSigHi] \\
\midrule
\multicolumn{5}{l}{\emph{2WikiMultiHopQA replication (pre-specified)}, $n=\WikiFictNhard$} \\
M0 single layer & \WikiFictMZeroNom & \WikiFictMZeroCe & \WikiFictMZeroEe & \WikiFictMZeroSig{} [\WikiFictMZeroSigLo, \WikiFictMZeroSigHi] \\
M0-last & \WikiFictMZerolastNom & \WikiFictMZerolastCe & \WikiFictMZerolastEe & \WikiFictMZerolastSig{} [\WikiFictMZerolastSigLo, \WikiFictMZerolastSigHi] \\
M1 layer mix & \WikiFictMOneNom & \WikiFictMOneCe & \WikiFictMOneEe & \WikiFictMOneSig{} [\WikiFictMOneSigLo, \WikiFictMOneSigHi] \\
M2 set pool & \WikiFictMTwoNom & \WikiFictMTwoCe & \WikiFictMTwoEe & \WikiFictMTwoSig{} [\WikiFictMTwoSigLo, \WikiFictMTwoSigHi] \\
M3 set attention & \WikiFictMThreeNom & \WikiFictMThreeCe & \WikiFictMThreeEe & \WikiFictMThreeSig{} [\WikiFictMThreeSigLo, \WikiFictMThreeSigHi] \\
M4-R readout+LoRA & \WikiFictMFoureNom & \WikiFictMFoureCe & \WikiFictMFoureEe & \WikiFictMFoureSig{} [\WikiFictMFoureSigLo, \WikiFictMFoureSigHi] \\
M4-J judge+LoRA & \WikiFictMFourjNom & \WikiFictMFourjCe & \WikiFictMFourjEe & \WikiFictMFourjSig{} [\WikiFictMFourjSigLo, \WikiFictMFourjSigHi] \\
Judge zero-shot & \WikiFictJudgeNom & \WikiFictJudgeCe & \WikiFictJudgeEe & \WikiFictJudgeSig{} [\WikiFictJudgeSigLo, \WikiFictJudgeSigHi] \\
\bottomrule
\end{tabular}
\caption{All ChainCheck metrics on closed-book-hard pairs (synthetic bridge entities). Nominal is AUC(A$>$D); chain effect is AUC(B$>$D)$-0.5$; edit effect is AUC(A$>$B)$-0.5$; selectivity is chain effect minus the absolute edit effect, with a 95\% percentile pair-bootstrap interval (2,000 resamples of accepted quads; not clustered by source question or document). For trained systems the point estimate is a median over three seeds, while the interval resamples seed-averaged per-quad outcomes, so it is not an interval for the displayed median. M0--M4 are trained on the nominal contrast; the judge is untrained.}
\label{tab:fict}
\end{table}

\begin{table}[!ht]
\centering\small
\begin{tabular}{lcccc}
\toprule
System & Nominal & Chain effect & Edit effect & $\Sigma$ seed median [seed-avg.\ boot.] \\
\midrule
\multicolumn{5}{l}{\emph{MuSiQue confirmation (pre-specified)}, $n=\MuRealNhard$} \\
Qwen3.8-27B (reference) & \MuRealJudgeNom & \MuRealJudgeCe & \MuRealJudgeEe & \MuRealJudgeSig{} [\MuRealJudgeSigLo, \MuRealJudgeSigHi] \\
Llama-3.1-8B-Instruct & \MuRealLlamaNom & \MuRealLlamaCe & \MuRealLlamaEe & \MuRealLlamaSig{} [\MuRealLlamaSigLo, \MuRealLlamaSigHi] \\
phi-4 & \MuRealPhiNom & \MuRealPhiCe & \MuRealPhiEe & \MuRealPhiSig{} [\MuRealPhiSigLo, \MuRealPhiSigHi] \\
Qwen3-32B & \MuRealQwenNom & \MuRealQwenCe & \MuRealQwenEe & \MuRealQwenSig{} [\MuRealQwenSigLo, \MuRealQwenSigHi] \\
Gemma-4-31B-it & \MuRealGemmaNom & \MuRealGemmaCe & \MuRealGemmaEe & \MuRealGemmaSig{} [\MuRealGemmaSigLo, \MuRealGemmaSigHi] \\
\midrule
\multicolumn{5}{l}{\emph{2WikiMultiHopQA replication (pre-specified)}, $n=\WikiRealNhard$} \\
Qwen3.8-27B (reference) & \WikiRealJudgeNom & \WikiRealJudgeCe & \WikiRealJudgeEe & \WikiRealJudgeSig{} [\WikiRealJudgeSigLo, \WikiRealJudgeSigHi] \\
Llama-3.1-8B-Instruct & \WikiRealLlamaNom & \WikiRealLlamaCe & \WikiRealLlamaEe & \WikiRealLlamaSig{} [\WikiRealLlamaSigLo, \WikiRealLlamaSigHi] \\
phi-4 & \WikiRealPhiNom & \WikiRealPhiCe & \WikiRealPhiEe & \WikiRealPhiSig{} [\WikiRealPhiSigLo, \WikiRealPhiSigHi] \\
Qwen3-32B & \WikiRealQwenNom & \WikiRealQwenCe & \WikiRealQwenEe & \WikiRealQwenSig{} [\WikiRealQwenSigLo, \WikiRealQwenSigHi] \\
Gemma-4-31B-it & \WikiRealGemmaNom & \WikiRealGemmaCe & \WikiRealGemmaEe & \WikiRealGemmaSig{} [\WikiRealGemmaSigLo, \WikiRealGemmaSigHi] \\
\bottomrule
\end{tabular}
\caption{Zero-shot judges from four model families, same prompt and score definition as the reference judge, real bridge entities. Every judge is reported; none is selected. Synthetic-entity results are in Table~\ref{tab:judges_fict}.}
\label{tab:judges}
\end{table}

\begin{table}[!ht]
\centering\small
\begin{tabular}{lcccc}
\toprule
System & Nominal & Chain effect & Edit effect & $\Sigma$ seed median [seed-avg.\ boot.] \\
\midrule
\multicolumn{5}{l}{\emph{MuSiQue confirmation (pre-specified)}, $n=\MuFictNhard$} \\
Qwen3.8-27B (reference) & \MuFictJudgeNom & \MuFictJudgeCe & \MuFictJudgeEe & \MuFictJudgeSig{} [\MuFictJudgeSigLo, \MuFictJudgeSigHi] \\
Llama-3.1-8B-Instruct & \MuFictLlamaNom & \MuFictLlamaCe & \MuFictLlamaEe & \MuFictLlamaSig{} [\MuFictLlamaSigLo, \MuFictLlamaSigHi] \\
phi-4 & \MuFictPhiNom & \MuFictPhiCe & \MuFictPhiEe & \MuFictPhiSig{} [\MuFictPhiSigLo, \MuFictPhiSigHi] \\
Qwen3-32B & \MuFictQwenNom & \MuFictQwenCe & \MuFictQwenEe & \MuFictQwenSig{} [\MuFictQwenSigLo, \MuFictQwenSigHi] \\
Gemma-4-31B-it & \MuFictGemmaNom & \MuFictGemmaCe & \MuFictGemmaEe & \MuFictGemmaSig{} [\MuFictGemmaSigLo, \MuFictGemmaSigHi] \\
\midrule
\multicolumn{5}{l}{\emph{2WikiMultiHopQA replication (pre-specified)}, $n=\WikiFictNhard$} \\
Qwen3.8-27B (reference) & \WikiFictJudgeNom & \WikiFictJudgeCe & \WikiFictJudgeEe & \WikiFictJudgeSig{} [\WikiFictJudgeSigLo, \WikiFictJudgeSigHi] \\
Llama-3.1-8B-Instruct & \WikiFictLlamaNom & \WikiFictLlamaCe & \WikiFictLlamaEe & \WikiFictLlamaSig{} [\WikiFictLlamaSigLo, \WikiFictLlamaSigHi] \\
phi-4 & \WikiFictPhiNom & \WikiFictPhiCe & \WikiFictPhiEe & \WikiFictPhiSig{} [\WikiFictPhiSigLo, \WikiFictPhiSigHi] \\
Qwen3-32B & \WikiFictQwenNom & \WikiFictQwenCe & \WikiFictQwenEe & \WikiFictQwenSig{} [\WikiFictQwenSigLo, \WikiFictQwenSigHi] \\
Gemma-4-31B-it & \WikiFictGemmaNom & \WikiFictGemmaCe & \WikiFictGemmaEe & \WikiFictGemmaSig{} [\WikiFictGemmaSigLo, \WikiFictGemmaSigHi] \\
\bottomrule
\end{tabular}
\caption{As Table~\ref{tab:judges}, synthetic bridge entities.}
\label{tab:judges_fict}
\end{table}

\begin{figure}[!htbp]
\centering
\includegraphics[width=\textwidth]{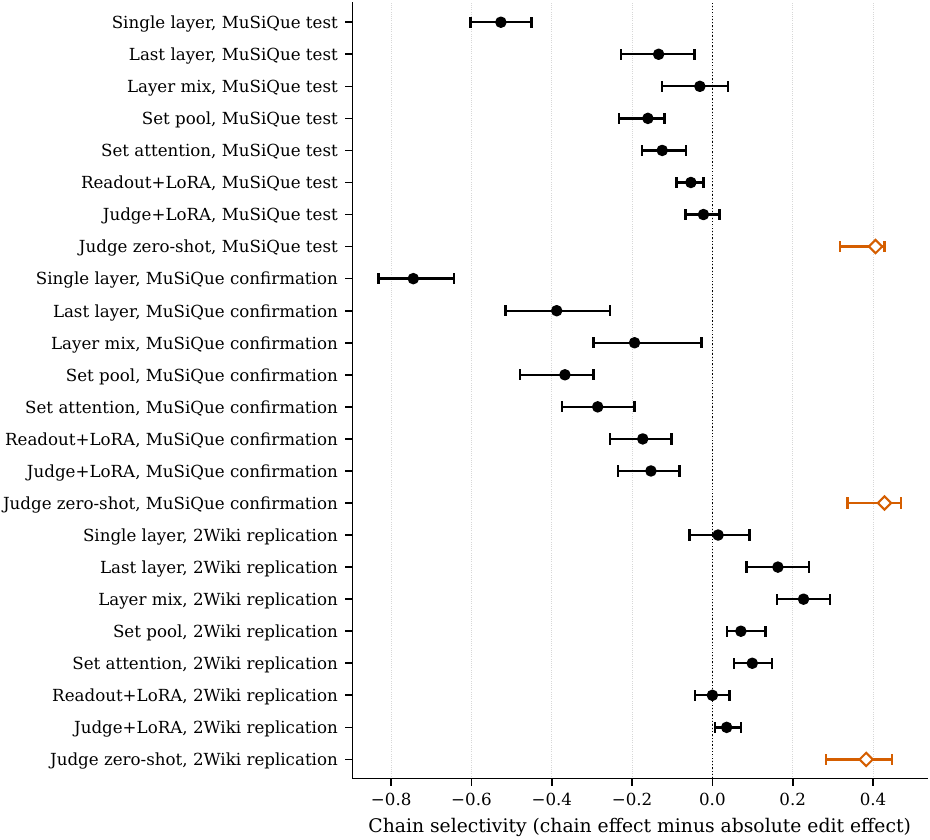}
\caption{Chain selectivity with 95\% pair-bootstrap intervals, synthetic bridge entities. Open diamonds
mark the reference zero-shot judge.}
\label{fig:fict}
\end{figure}

\begin{figure}[!htbp]
\centering
\includegraphics[width=\textwidth]{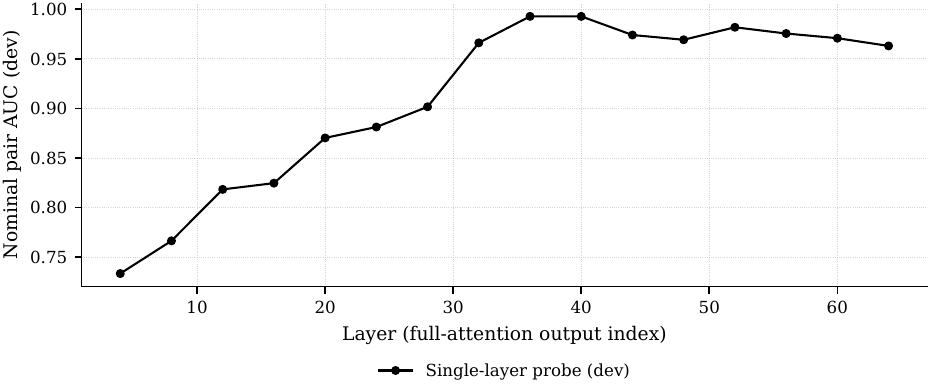}
\caption{Nominal dev pair AUC of a single-layer probe on each full-attention layer of the backbone.}
\label{fig:layers}
\end{figure}

\end{document}